\documentclass{article}

\usepackage[final]{neurips_2025}
\usepackage[utf8]{inputenc} 
\usepackage[T1]{fontenc}    
\usepackage{hyperref}       
\usepackage{url}            
\usepackage{booktabs}       
\usepackage{amsfonts}       
\usepackage{siunitx} 
\usepackage{nicefrac}       
\usepackage{microtype}      
\usepackage{nicematrix} 
\usepackage{xcolor}         
\usepackage{amsmath}
\usepackage{amssymb}
\usepackage{mathtools}
\usepackage{amsthm}
\usepackage{graphicx}
\usepackage{wrapfig}
\usepackage{lipsum}
\usepackage{color, colortbl}
\usepackage{bbm}
\usepackage{subcaption}
\usepackage{multirow}
\usepackage{tikz}
\usepackage{enumitem}
\usepackage{xspace}
\usepackage[most]{tcolorbox}
\tcbuselibrary{listingsutf8}
\usepackage{makecell}
\usepackage{arydshln}
\newcolumntype{g}{>{\cellcolor{Gray}}c}
\usepackage{subcaption}

\definecolor{tableofcontent}{HTML}{E63E15}
\definecolor{urlcol}{HTML}{2470D8}
\usepackage[normalem]{ulem}
\useunder{\uline}{\ul}{}
\hypersetup{
    colorlinks=true,       
    linkcolor=tableofcontent, 
    urlcolor=black,           
}

\newcommand{\pgftextcircled}[1]{
    \setbox0=\hbox{#1}%
    \dimen0\wd0%
    \divide\dimen0 by 2%
    \begin{tikzpicture}[baseline=(a.base)]%
        \useasboundingbox (-\the\dimen0,0pt) rectangle (\the\dimen0,1pt);
        \node[circle,draw,outer sep=0pt,inner sep=0.1ex] (a) {#1};
    \end{tikzpicture}
}
\setlist{leftmargin=10mm}
\definecolor{Gray}{gray}{0.9}
\newcommand{\xhdr}[1]{{\vspace{1pt}\noindent\bfseries #1}.}
\newcommand{\ie}{\textit{i.e., }}
\newcommand{\eg}{\textit{e.g., }}

\usepackage[most]{tcolorbox}
\usepackage{lipsum}
\tcbuselibrary{skins}
\usepackage{mdframed}
\definecolor{niceblue}{HTML}{3c9dfd}
\usepackage{fontawesome5}
\usepackage{pifont}
\newcommand{\cmark}{\ding{51}}%
\newcommand{\xmark}{\ding{55}}%

\newtcolorbox{Mycolorbox}[2][]{
  arc=5mm,
  lower separated=false,
  fonttitle=\bfseries,
  colbacktitle=gray,
  coltitle=white,
  enhanced,
  attach boxed title to top left={xshift=0.5cm, yshift=-2mm},
  colframe=gray,
  colback=white,
  title=#1, #2,
  breakable
}

\newcommand{\name}{\texttt{TRACE}\xspace}
\usepackage[capitalize,noabbrev]{cleveref}

\theoremstyle{plain}

\theoremstyle{definition}

\theoremstyle{remark}

\title{\name: Grounding Time Series in Context for Multimodal Embedding and Retrieval}

\author{
  Jialin Chen\textsuperscript{1\thanks{Both authors contributed equally to this paper.}}, 
  Ziyu Zhao\textsuperscript{2\footnotemark[1]}, 
  Gaukhar Nurbek\textsuperscript{3}, 
  Aosong Feng\textsuperscript{1}, \\
  \textbf{Ali Maatouk\textsuperscript{1}, 
  Leandros Tassiulas\textsuperscript{1}, Yifeng Gao\textsuperscript{3}, Rex Ying\textsuperscript{1}} \\
  \textsuperscript{1}Yale University, 
  \textsuperscript{2}McGill University, 
  \textsuperscript{3}University of Texas Rio Grande Valley \\
  \texttt{\{jialin.chen, aosong.feng, ali.maatouk, leandros.tassiulas, rex.ying\}@yale.edu}, \\
  \texttt{ziyu.zhao2@mail.mcgill.ca}; \texttt{\{gaukhar.nurbek01, yifeng.gao\}@utrgv.edu} \\
}

\begin{document}

\maketitle

\begin{abstract}
The ubiquity of dynamic data in domains such as weather, healthcare, and energy underscores a growing need for effective interpretation and retrieval of time-series data. These data are inherently tied to domain-specific contexts, such as clinical notes or weather narratives, making cross-modal retrieval essential not only for downstream tasks but also for developing robust time-series foundation models by retrieval-augmented generation (RAG). Despite the increasing demand, time-series retrieval remains largely underexplored. Existing methods often lack semantic grounding, struggle to align heterogeneous modalities, and have limited capacity for handling multi-channel signals. To address this gap, we propose \name, a generic multimodal retriever that grounds time-series embeddings in aligned textual context. \name enables fine-grained channel-level alignment and employs hard negative mining to facilitate semantically meaningful retrieval. It supports flexible cross-modal retrieval modes, including Text-to-Timeseries and Timeseries-to-Text, effectively linking linguistic descriptions with complex temporal patterns. By retrieving semantically relevant pairs, \name enriches downstream models with informative context, leading to improved predictive accuracy and interpretability. Beyond a static retrieval engine, \name also serves as a powerful standalone encoder, with lightweight task-specific tuning that refines context-aware representations while maintaining strong cross-modal alignment. These representations achieve state-of-the-art performance on downstream forecasting and classification tasks. Extensive experiments across multiple domains highlight its dual utility, as both an effective encoder for downstream applications and a general-purpose retriever to enhance time-series models \footnote{Codes are available at \url{https://github.com/Graph-and-Geometric-Learning/TRACE-Multimodal-TSEncoder}.}.
\end{abstract}

\section{Introduction}
Time-series data is prevalent across critical domains such as healthcare, weather, and energy~\cite{energy_1,energy_2,Weather,Graphcast}. Crucially, such data rarely exists in isolation in real-world applications. It is typically accompanied by rich, domain-specific textual context, \eg clinical notes and weather reports~\cite{liu2024time,chen2025mtbenchmultimodaltimeseries,lee2025timecaplearningcontextualizeaugment}. This inherent multimodality necessitates a shift beyond unimodal time-series analysis towards multi-modal frameworks that seamlessly integrate these heterogeneous data types. 

\begin{figure}[t]
    \centering
    \includegraphics[width=\linewidth]{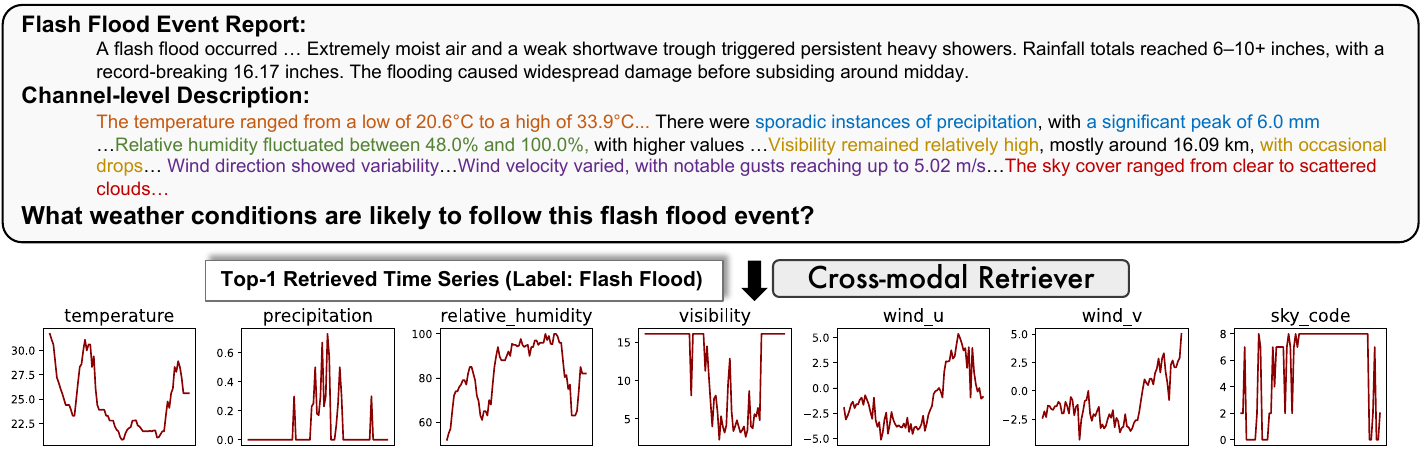}\vspace{-0.3cm}
    \caption{A Use Case of Text-to-Timeseries Retrieval}
    \label{fig:usecase}\vspace{-0.7cm}
\end{figure}

Cross-modal retrieval between time series and text is not only natural but necessary. As shown in Figure~\ref{fig:usecase}, given a flash flood report describing extreme rainfall and high wind gusts, retrieving historical time series that exhibit similar patterns can support downstream tasks such as weather forecasting and disaster warning.  Such retrieval also enables the integration of semantically aligned external knowledge into time series foundation models~\cite{goswami2024moment, liu2024timer, shi2024time}, guiding model attention to relevant segments, and facilitating more generalizable inference via retrieval-augmented generation (RAG).

Despite the clear demand, time-series retrieval, particularly in a cross-modal context, remains significantly underexplored. Existing approaches often fall short in several ways~\cite{yeh2023efficient, zhang2024timeraf, liu2024retrieval,ning2025ts,yang2025timerag}. They overlook the rich textual context within time-series data and rely on shallow similarity measures rather than contextual understanding, leading to a lack of effective cross-modal alignment between time-series signals and their associated textual descriptions. Moreover, they struggle with the multi-channel nature of real-world time series, where each channel can encode distinct yet interrelated information~\cite{CICD, kim2025comprehensive, chen2024similarity}. Importantly, prior work rarely explores retrieval-augmented generation (RAG) for time series foundation models, restricting their utility in augmenting downstream models. 

\begin{figure}[htbp]
    \vspace{-0.5cm}
    \centering
    \includegraphics[width=\linewidth]{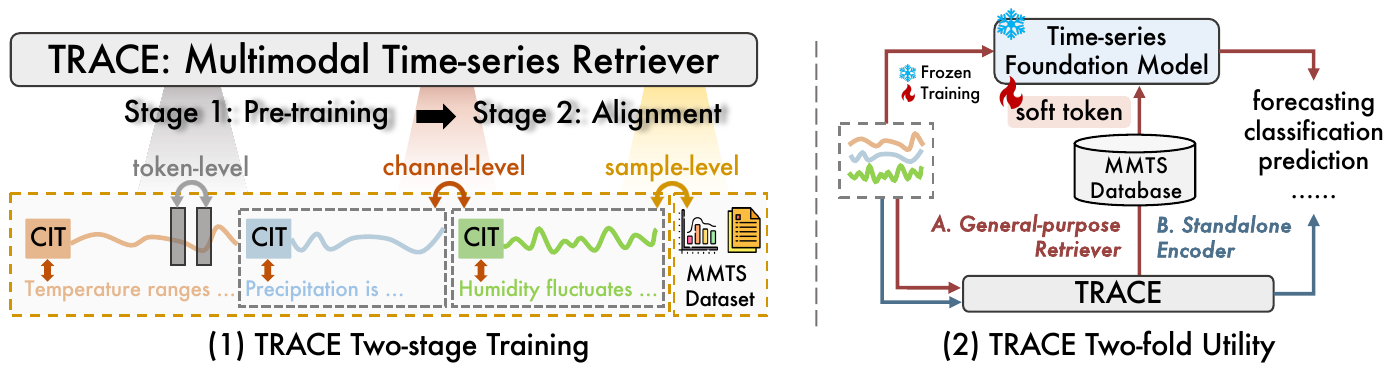}\vspace{-0.2cm}
    \caption{Overview of \name. CIT stands for Channel Identity Tokens, which serve as a key bridge to connect two stages. MMTS denotes multimodal time series.}
    \label{fig:cover_fig}\vspace{-0.3cm}
\end{figure}

To address this gap, we introduce \name, a novel multimodal \textbf{T}ime-series \textbf{R}etriever with \textbf{A}ligned \textbf{C}ontext \textbf{E}mbedding. As illustrated in Figure~\ref{fig:cover_fig}, \name adopts a two-stage training: a pre-training stage for the time-series encoder, followed by a cross-modal alignment. To address the challenge of modeling multivariate time series, we introduce Channel Identity Tokens (CITs) into a masked autoencoder framework pre-trained at both the token level and channel level in Stage 1. CITs guide the model to attend to unique channel behaviors and enable the learning of channel disentangled representations, overcoming the limitation of conventional decoder-only foundation models which often yield embeddings lacking discriminative power for retrieval and classification. In Stage 2, we propose a novel component for effective cross-modal alignment between time-series embeddings and their textual counterparts through a hierarchical hard negative mining strategy. At the channel level, we identify distractor single-channel segments that exhibit misleadingly similar patterns. At the sample level, we dynamically mine hard negatives by selecting highly similar text descriptions but with divergent semantics. This dual-level contrastive learning encourages the model to learn both local precision and global consistency, leading to strong generalization in downstream tasks.

\name is designed with a two-fold utility. It acts as a general-purpose retriever, which provides relevant information via a soft token interface. The soft token summarizes retrieved time-series snippets into a latent vector, which is then prepended as a conditioning token, guiding a frozen time-series foundation model towards more context-aware predictions. Moreover, \name serves as a powerful standalone encoder, producing rich embeddings that achieve state-of-the-art performance on downstream forecasting and classification tasks. Extensive experiments on both public benchmarks and our curated multimodal dataset validate the effectiveness of \name, demonstrating superior retrieval accuracy. The retrieved context substantially boosts downstream time-series models in retrieval-augmented settings, with up to 4.56\% increase in classification accuracy and 4.55\% reduction in forecasting error. In addition, \name produces high-quality time-series embeddings that achieve state-of-the-art results on a wide range of forecasting and classification benchmarks.

The contributions of this paper are: (1) we propose the first multimodal retriever, \name, that learns semantically grounded time-series embeddings through fine-grained dual-level alignment; (2) we establish new benchmarks on cross-modal retrieval between time series and text, and (3) extensive validation showcases that \name consistently delivers state-of-the-art performance both as a general-purpose retriever for time-series models and a powerful encoder for time series analysis.
\section{Related Work}

\xhdr{Time Series Forecasting} Recent work on time-series forecasting has led to a range of model architectures, each emphasizing different inductive biases. Transformer-based models leverage self-attention to capture long-range dependencies and flexible temporal dynamics~\cite{wang2020linformer, liu2022non, Autoformer, Yuqietal-2023-PatchTST, zhang2022crossformer, iTransformer, probabilistic_transformer, Fedformer,Pyraformer,feng2024efficient}. Linear-based models assume time-series signals can be effectively decomposed and modeled with simple linear projections~\cite{DLinear,chen2023tsmixer}. Frequency-domain and mixing-based approaches aim to model periodicity and multi-scale temporal structures using Fourier transforms or token mixers~\cite{wang2024timemixer}. Recently, a variety of time series foundation models have emerged. Timer-XL~\cite{liu2024timer} leverages Kronecker attention and is pre-trained with multivariate next-token prediction to enable unified, long-context forecasting. Chronos~\cite{ansari2024chronos} tokenizes time series via scaling and quantization, and trains a T5-style model for zero-shot probabilistic forecasting. Time-MoE~\cite{shi2024time} introduces a sparse mixture-of-experts architecture to support variable horizons and input lengths. TimesFM~\cite{das2024decoder} uses input patching and is pre-trained on large-scale data for strong zero-shot performance. Moment~\cite{goswami2024moment} and Moirai~\cite{liu2024moirai} adopt masked prediction pretraining to enable generalization across diverse multivariate forecasting tasks. While these models perform well on forecasting tasks, they are generally unimodal and not designed for retrieval or integration of external context, highlighting a gap addressed by our cross-modal retrieval framework. 

\xhdr{Time Series Language Models} Recently, several multimodal encoders have been proposed to integrate time series and text~\cite{jin2023time, zhou2023one, hu2025context, liu2024timecma, pan2024s,zhao2025enhancing}, which aim to leverage the generalization capabilities of large language models by reprogramming time series into token-like representations or textual prototypes. ChatTime\cite{wang2025chattime} models time series as a foreign language by normalizing and discretizing continuous signals into token sequences, which are then processed by a large language model (LLM). ChatTS\cite{xie2024chatts} supports both understanding and reasoning by fine-tuning on synthetic datasets generated via attribute-based sampling. TimeXL\cite{jiang2025explainable} combines a prototype-based time series encoder with a multimodal prediction framework to capture explainable temporal patterns guided by aligned textual cues. However, they primarily treat text as global context and lack fine-grained alignment between structured time series components and textual semantics, leading to suboptimal cross-modal embedding or retrieval.

\xhdr{Time Series Retrieval System} Recent work has explored retrieval systems for time series data, primarily within a unimodal setting~\cite{yue2022ts2vec,liu2024retrieval,jing2022retrieval,yang2025timerag}. CTSR~\cite{yeh2023efficient} supports content-based time-series retrieval using contextual metadata. TimeRAF~\cite{zhang2024timeraf} integrates a trainable retriever with task-specific time-series knowledge bases for downstream augmentation. TS-RAG~\cite{ning2025ts} retrieves relevant time series segments using pre-trained encoders and combines them via a mixture-of-experts module to improve forecasting. However, all of these methods rely solely on time series embeddings and do not incorporate textual signals, limiting their ability to support multimodal and context-aware retrieval.

\section{Proposed Method}
\label{sec:method}
\begin{figure}
    \centering
    \includegraphics[width=\linewidth]{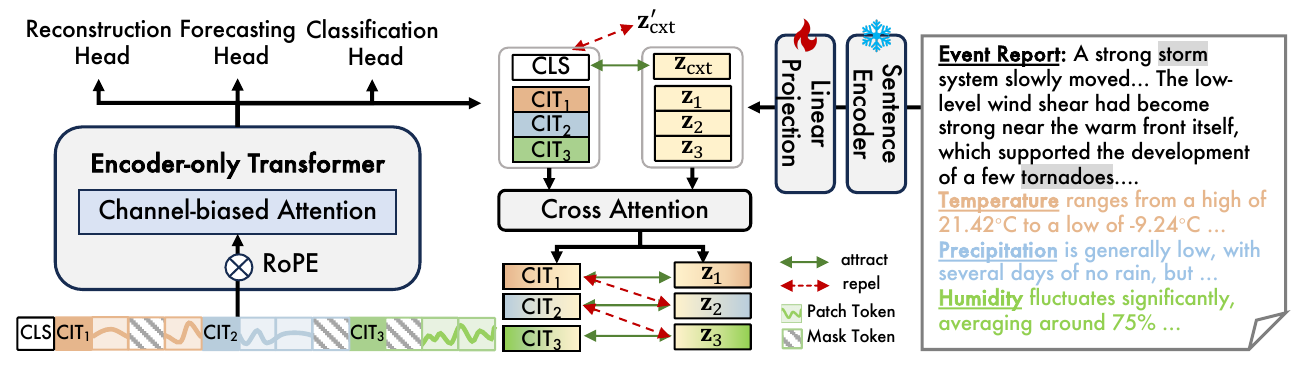}
    \caption{Illustration of \name, which encodes multivariate time series using channel-biased attention and aligns token embeddings with its corresponding textual description (\eg $\mathbf{z}_i$ and $\mathbf{z}_{\text{cxt}}$) through cross-attention and dual-level contrastive learning. $\mathbf{z}'_{\text{cxt}}$ indicates an in-batch hard negative sample.}
    \label{fig:architecture}\vspace{-0.5cm}
\end{figure}

As shown in Figure~\ref{fig:architecture}, \name learns robust time series representations through a masked reconstruction objective with channel-biased attention in the pre-training stage (Sec.~\ref{sec:stage1}). Then, each time series channel is aligned with its corresponding textual description via fine-grained contrastive learning in the cross-modal alignment stage (Sec.~\ref{sec:stage2}). We further propose a novel retrieval-augmented generation strategy for time series foundation models, where \name retrieves relevant context for downstream tasks (Sec.~\ref{sec:method_rag}). This modular design enables both strong standalone performance and effective integration with existing time series foundation models.
\subsection{Problem Definition}
\xhdr{Multimodal Time-series} Let $\mathbf{X} \in \mathbb{R}^{C \times T}$ denote a multivariate time series instance, where $C$ is the number of channels (or variables) and $T$ is the number of time steps. We assume the availability of two types of textual information aligned with $\mathbf{X}$. First, for each channel $c$ in an instance $\mathbf{X}$, there is a corresponding textual description $\tau_c$ that summarizes the behavior or trend of $\mathbf{X}_c$ over the time window $[0, T)$. These descriptions are denoted as $\mathcal{T}^{\text{ch}} = \{\tau_c|c=1,\cdots,C\}$.  Additionally, there is a sample-level context $\tau_{\text{cxt}}$ summarizing the overall condition occurring during the same time window, which could be weather reports or clinical narratives, depending on the application domain. 

\xhdr{Task Objectives} The goal is to jointly embed the multivariate time series $\mathbf{X}$ and its corresponding textual context $\mathcal{T} = \mathcal{T}^{\text{ch}} \cup \{\tau_{\text{cxt}}\}$ into a shared space that supports multiple downstream tasks, including: (1) forecasting future values $\mathbf{X}_{T:T+H} \in \mathbb{R}^{C \times H}$ for the next $H$ time steps; (2) classification, where the model predicts a categorical label for each time series instance; and (3) cross-modal retrieval, where the goal is to retrieve relevant time series $\mathbf{X}$ based on a text query $\tau_{\text{cxt}}$ or retrieve historical relevant reports from $\mathcal{T}$ given a time series query, etc.

\subsection{Stage 1: Time Series Encoder Pre-training}\label{sec:stage1}
\xhdr{Time Series Tokenization}
Given an input multivariate time series $\mathbf{X} \in \mathbb{R}^{C \times T}$, we divide the temporal dimension into non-overlapping (or strided) patches of length $P$, resulting in $\hat{T} = \lfloor \frac{T}{P} \rfloor$ patches per channel. Each patch is flattened and linearly projected into a $d$-dimensional embedding space using a learnable linear projection. This converts each channel into a sequence of patch tokens $X_c^{\text {patch }} \in \mathbb{R}^{\hat{T} \times d}$, for $\forall c \in\{1, \ldots, C\}$. To capture localized semantics within each channel, we prepend a learnable channel identity token [\texttt{CIT}] $\in \mathbb{R}^{1 \times d}$ to the patch token sequence of each channel. These tokens serve as explicit representations of channel-level summaries. Each token is uniquely indexed and not shared across channels, initialized from a standard Gaussian distribution, and trained jointly with the model. This design allows the model to differentiate between channels and effectively aggregate channel-wise patterns. We then concatenate all tokenized channels into a single sequence and insert a global learnable \texttt{[CLS]} token at the beginning of the full sequence. The final token sequence for a multivariate instance is structured as: 
\begin{equation}\vspace{-0.1cm}
    \mathbf{H}=\left[\texttt{[CLS]}; \texttt{[CIT]}_1 ; X_1^{\mathrm{patch}} ; \texttt{[CIT]}_2 ; X_2^{\mathrm{patch}} ; \ldots ; \texttt{[CIT]}_C ; X_C^{\mathrm{patch}}\right] \in \mathbb{R}^{L \times d},
    \label{eq:seq}
\end{equation}
where $L = C(\hat{T} + 1) + 1$ is the total sequence length after flattening all channel in \ref{eq:seq}. This tokenization strategy preserves both temporal and structural granularity: patchification encodes token-level patterns; \texttt{[CIT]} summarizes intra-channel dynamics; and \texttt{[CLS]} provides a global and sample-level embedding that can be used for downstream retrieval and classification tasks. 

\xhdr{Channel-biased Attention and Rotary PE}
To encode channel dependencies in multivariate time series, we introduce a novel Channel-biased Attention (\texttt{CbA}) mechanism that incorporates both inductive bias for channel disentanglement and temporal order encoding via rotary positional embeddings (RoPE)~\cite{su2024roformer}.
In our \texttt{CbA}, we design a biased attention mask $M \in \{0, 1\}^{L \times L}$ to prevent unintended semantic entanglement across heterogeneous variables. Specifically, for each channel identity token \texttt{[CIT]}$_c$ located at index $i_c$ in the flattened sequence, we define $M_{i_c,j}=0$ if token $j\notin$ channel $c$ and 1 otherwise, and $M_{k,j}=1$ if token $k$ is not a \texttt{[CIT]}. Let $\mathbf{Q}, \mathbf{K}, \mathbf{V} \in \mathbb{R}^{L \times d}$ be the learned linear projections of the input token embedding $\mathbf{H}$. 
We apply RoPE to the query ($\mathbf{Q}$) and key ($\mathbf{K}$) vectors before computing attention. RoPE is applied independently within each channel to the $\hat{T}$ temporal tokens, and is not applied to the channel identity tokens, which act as position-agnostic aggregators. The attention weight between tokens $i$ and $j$ in a RoPE-enhanced attention is given by $\alpha_{i j}=\operatorname{softmax}_j\left({Q_i^{\top} R_{\theta_{\Delta t_{i j}}} K_j}/{\sqrt{d}}+\log M_{i j}\right)$, where $R_{\theta_{\Delta t_{ij}}}(\cdot)$ denotes a rotation by angle $\theta_{\Delta t_{ij}}$, and $\Delta t_{ij}$ is the relative time difference between tokens $i$ and $j$ in their original unflattened sequence. This is crucial in the multichannel setting, as two tokens that are close in actual time may appear far apart in the flattened sequence. Using $\Delta t_{ij}$ ensures that the position encoding remains consistent with the true temporal structure rather than the flattened channel order. $M_{i j}$ mask enforces channel disentanglement, while still allowing rich token-level interactions across the full sequence.

\xhdr{Pre-training Setup}
We adopt an encoder-only Transformer~\cite{vaswani2017attention} with multi-head channel-based attention layers in \name. We apply reversible instance normalization~\cite{kim2021reversible} to multivariate time series before tokenizing and embedding. A fixed proportion of these tokens is randomly masked with a mask ratio of $\gamma$, and the model is pre-trained to reconstruct the missing values based on the unmasked context. We use mean squared error (MSE) loss to supervise pre-training, encouraging the model to capture cross-channel dependencies while learning transferable representations for downstream tasks.

\subsection{Stage 2: Multimodal Alignment Learning}\label{sec:stage2}
\xhdr{Motivation} Standard contrastive learning methods typically rely on sample-level random negatives. However, textual descriptions frequently reference specific variables (\eg temperature spikes, wind gusts), which cannot be precisely aligned using a single global embedding. To address this, we introduce channel-level alignment that explicitly models the interaction between individual time-series channels and their corresponding textual context. This not only enhances semantic precision but also promotes modularity in representation learning and enables variable-specific interactions.

\xhdr{Cross-attention Between Modalities}
After pre-training the time-series encoder via masked reconstruction, we obtain hidden embedding $\mathbf{H}^\text{out} \in \mathbb{R}^{L \times d}$ from the final transformer layer, where $L$ is the full sequence length after flattening all channels. From this, we extract the \texttt{[CLS]} token embedding $\mathbf{h}_{\texttt{[CLS]}} \in \mathbb{R}^d$, and the set of channel identity token embeddings $\mathbf{H}_{\texttt{[CIT]}} = [\mathbf{h}_1, \dots, \mathbf{h}_C] \in \mathbb{R}^{C \times d}$, each corresponding to a \texttt{[CIT]} token and serving as fine-grained anchors that enable structured reasoning at the channel level. Let $\tau_{\text{cxt}}$ and $\tau_c$ denote the sample-level and the $c$-th channel textual context for a time series instance, respectively. The textual inputs are first encoded using a pre-trained language model (\eg a frozen Sentence-Transformer~\cite{reimers-2019-sentence-bert}), followed by a learnable linear layer that projects them into the same $d$-dimensional embedding space as the time series representations, collectively denoted as $f_{\text{t}}(\cdot)$. This yields semantic embeddings $\mathbf{z}_{\mathrm{cxt}} = f_{\text{t}}(\tau_{\mathrm{cxt}}) \in \mathbb{R}^d$ for the sample-level context and $\mathbf{z}_c = f_{\text{t}}(\tau_c) \in \mathbb{R}^d$ for each channel-level description.  We further apply a cross-attention between $\mathbf{H}_{\texttt{[CIT]}} \in \mathbb{R}^{C \times d}$ and channel text embeddings $\mathbf{Z}_{\text{ch}} = [\mathbf{z}_1, \ldots, \mathbf{z}_C] \in \mathbb{R}^{C \times d}$, allowing information to be fused across aligned channels. This interaction allows the model to refine its channel-wise time-series representations using semantically aligned textual information.

\xhdr{Dual-level Hard Negative Mining} To enhance the discriminative capacity of the model, we develop a dual-level hard negative mining strategy that introduces fine-grained contrastive pressure at both the sample and channel levels. This approach enables the model to distinguish not only between unrelated time series and text, but also between subtly confusable pairs that share superficial temporal similarity but diverge semantically.  For each time series instance $i$, we mine negative candidates from all other sample-level reports in the same batch based on embedding cosine similarity. For a certain channel, we mine channel-level negatives from a broader candidate pool that includes both intra-instance distractors (other channels within the same sample) and inter-instance distractors (same-indexed channels across different samples). Specifically, for the $c$-th channel of the $i$-th instance, we define the sample-level and channel-level negative candidate set as \vspace{-0.2cm}
\begin{equation*}\vspace{-0.1cm}
    \mathcal{N}_{\mathrm{cxt}}^{(i)}=\operatorname{Top}_K\left\{\operatorname{sim}(\mathbf{h}_{\texttt{[CLS]}}^{(i)}, \mathbf{z}_{\mathrm{cxt}}^{(j)}) \mid j \neq i\right\}, \mathcal{N}_{\mathrm{ch}}^{(i, c)}=\operatorname{Top}_K\left\{\operatorname{sim}(\mathbf{h}_c^{(i)}, \mathbf{z}_{c^{\prime}}^{(j)}) \mid c^{\prime} \neq c \text{ or } j\neq i\right\},
\end{equation*} 
where $K$ is number of negative samples at each level. Symmetric negative sets are defined in the reverse direction for $\mathbf{z}_{\mathrm{cxt}}^{(i)}$ and  $\mathbf{z}_{\mathrm{c}}^{(i)}$ by swapping the roles of time series and text. We then compute a bidirectional InfoNCE loss at sample levels: $\mathcal{L}_{\mathrm{global}}^{\text{text} \rightarrow \text{ts}}$, $\mathcal{L}_{\mathrm{global}}^{\text{ts} \rightarrow \text{text}}$, and similarly for channel-level losses. The total alignment objective is the average of both directions (Formulations detailed in Appendix~\ref{app:method_details}): \vspace{-0.3cm}
\begin{equation}\vspace{-0.1cm}
    \mathcal{L}_{\text {align }}=\frac{1}{2}\left(\mathcal{L}_{\text {global }}^{\text {text }\rightarrow\text { ts }}+\mathcal{L}_{\text {global }}^{\text {ts }\rightarrow\text { text }}\right)+\lambda_{\text {ch }} \cdot \frac{1}{2}\left(\mathcal{L}_{\text {channel }}^{\text {text }\rightarrow\text { ts }}+\mathcal{L}_{\text {channel }}^{\text {ts }\rightarrow\text { text }}\right),
\end{equation}
 where $\lambda_{\text{ch}}$ controls the contribution of channel-level alignment. The entire alignment objective is optimized jointly with the trainable parameters of the time series encoder in the pre-training stage and the linear projection head in $f_{\text{t}}$, while keeping the backbone language model frozen.

\subsection{Retrieval-augmented Generation with Time Series Foundation Models}\label{sec:method_rag}
As shown in Figure~\ref{fig:cover_fig}, \name enables retrieval-augmented generation (RAG) for time series foundation models, inspired by the success of RAG in NLP~\cite{gao2023retrieval,liu2024retrieval}. Given a query time series, \name computes its \texttt{[CLS]} token embedding and retrieves the top-$R$ most relevant multimodal pairs ${(\mathbf{X}^i, \tau_{\text{cxt}}^i)}_{i=1}^R$ from the pre-built multimodal database based on the embedding similarity, where $\mathbf{X}^i$ is a historical multivariate time series and $\tau_{\text{cxt}}^i$ is the associated sample-level context. Specifically, the time series component is encoded to $\mathbf{h}_{\text{ts}}^{(i)} \in \mathbb{R}^d$, and the textual context $\tau_{\text{cxt}}^i$ is encoded to $\mathbf{z}_{\text{cxt}}^{(i)} \in \mathbb{R}^d$ (as in Sec.~\ref{sec:stage2}). These representations are concatenated, stacked, and mapped through a single trainable projection layer to generate a final, dense soft token $\mathbf{P}$, which serves as a continuous prompt that is prepended to the query sequence input. This design allows the downstream forecaster to incorporate external knowledge without architectural modification. Importantly, the base time-series foundation model remains frozen during training; only the projection layer and a lightweight task-specific head are updated. This approach ensures efficiency and model-agnosticism, enabling plug-and-play integration across diverse backbone architectures. In effect, \name acts as a structured, external memory, enriching the model's input with historically grounded and semantically aligned context.



\section{Experiments}
\label{sec:experiments}
We evaluate \name from three key perspectives: (1) its effectiveness in cross-modal retrieval (Sec.~\ref{sec:retrieval}) and time-sereis retrieval (Sec.~\ref{sec:ts2ts}) compared to strong baselines, (2) its utility as a retriever in retrieval-augmented forecasting pipelines (Sec~\ref{sec:rag}), and (3) its generalization ability as a standalone encoder for forecasting and classification (Sec.~\ref{sec:encoder}). Experiments are conducted on public benchmarks and our curated multimodal dataset designed to assess cross-modal alignment and retrieval performance.

\subsection{Experimental Setting}
\xhdr{Dataset}
To support real-world multimodal time series applications, we construct a new dataset in the weather domain with three aligned components: multivariate time series, sample-level event reports, and synthetic channel-level descriptions, specifically for downstream forecasting and event-type classification tasks. The event reports are sourced from the NOAA Events Database~\cite{stormevents}, while the associated time series data are retrieved from the NOAA Global Historical Climatology Network (GHCN)~\cite{ghcnhourly2023}. We focus on stations and time windows characterized by frequent severe weather events and extract historical multivariate time-series segments at multiple temporal resolutions, anchored at event onset. To enhance data diversity and model robustness, we also sample non-event (\ie typical) periods from the same stations, as well as from geographically distinct locations. Each time-series segment includes seven variables (\eg temperature, relative humidity, precipitation) and is annotated with either a specific event type or a non-event label. To evaluate performance in the univariate setting, we further incorporate the three largest subsets—Health, Energy, and Environment—from TimeMMD~\cite{liu2024time}, a multimodal benchmark designed for time series forecasting, where each single-variate instance is aligned with a sample-level textual report (\eg clinical notes, incident logs). This setting allows us to assess the model’s generalization across diverse domains and varying channel configurations. Full dataset details and illustrative examples are provided in Appendix~\ref{app:our_dataset}.

\xhdr{Baselines}
We evaluate against the state-of-the-art traditional time series models and recent time series foundation models. Traditional baselines include DLinear~\cite{DLinear}, iTransformer~\cite{iTransformer}, PatchTST~\cite{Yuqietal-2023-PatchTST}, TimesNet~\cite{wu2023timesnet}, TimeMixer~\cite{wang2024timemixerdecomposablemultiscalemixing}, and multimodal model FSCA~\cite{hu2025context}. These models are trained from scratch on each task. For foundation models, we include Chronos\cite{ansari2024chronos}, TimesFM~\cite{das2024decoder}, Timer-XL~\cite{liu2024timer}, Time-MoE~\cite{shi2024time}, Moirai~\cite{liu2024moirai} and Moment~\cite{goswami2024moment}. We refer to Appendix~\ref{app:baseline} for baseline details.

\xhdr{Implementation Details}
The default \name consists of a 6-layer Transformer encoder with a hidden dimension of $384$ and $6$ attention heads. We use the AdamW~\cite{loshchilov2017decoupled} optimizer with a linear warmup followed by a cosine decay schedule. Pre-training is conducted with a mask ratio of $0.3$, and runs for up to $400$ epochs. We take $32$ in-batch negative samples at each level in the alignment stage and run for up to $300$ epochs. All experiments are conducted over five runs with different random seeds on NVIDIA A100 40GB GPUs. We refer to Appendix~\ref{app:hyper} for experiment configurations and details. 

\subsection{Cross-modal Retrieval}\label{sec:retrieval}
\xhdr{Alignment Setup} To evaluate the model's retrieval performance, we conduct a controlled comparison by replacing the encoder in \name with several strong time series foundation models that produce fixed-length embeddings. Each encoder is jointly fine-tuned end-to-end with a lightweight projection layer following the sentence encoder, using a contrastive learning objective. While \name leverages \texttt{[CLS]} and \texttt{[CIT]} embeddings for dual-level alignment, other baselines use mean pooling over the sequence due to their architectural constraints.

\xhdr{Evaluation Metrics}
\name supports flexible retrieval modes, including cross-modal (Text-to-TS and TS-to-Text) and unimodal TS-to-TS retrieval. For cross-modal retrieval, a query in one modality is used to retrieve its corresponding counterpart in the other modality based on embedding cosine similarity. The evaluation includes several metrics:
\begin{itemize}[leftmargin=2mm,itemsep=0.1pt,topsep=0.3pt]
    \item \textbf{Label Matching} uses P@$k$ to measure the precision of correctly labeled items among the top-$k$ retrieval, and Mean Reciprocal Rank (MRR) to assess the rank of the first correct item.
    \item \textbf{Modality Matching} evaluates whether a query retrieves its paired instance from the opposite modality, using P@$k$ for top-$k$ precision and MRR for the rank of the true counterpart.
    \item \textbf{Text Similarity} uses ROUGE between the query text and the text paired with the top-1 retrieved time series (for text-to-ts scenario), or between the top-1 retrieved text and the original text paired with the query time series (for ts-to-text scenario).
    \item \textbf{Time Series Similarity} computes MAE and MSE between the time series linked to the query and that of the top-1 retrieved pair, defined similarly to Text Similarity.
\end{itemize}
\xhdr{Results}
As shown in Table~\ref{tab:retrieval}, \name consistently achieves state-of-the-art performance in two retrieval settings with approximately 90\% top-1 label matching and 44\% top-1 modality matching. Notably, this retrieval precision surpasses the classification accuracy of all train-from-scratch models reported in Table~\ref{tab:classification}, highlighting the strength of alignment supervision in learning discriminative representations. Among baselines, Moment outperforms other foundation models, suggesting that encoder-only architectures are better suited for dense retrieval tasks. In contrast, \name provides fine-grained embeddings for cross-modal alignment, enabling it to recover semantical counterparts with high precision. 
\begin{table}[h]\vspace{-0.3cm}
\centering
\caption{Retrieval results on 2,000 bidirectional Text–Timeseries query pairs. ``Random'' indicates a non-informative retriever that ranks candidates uniformly at random.}\label{tab:retrieval}
\resizebox{\linewidth}{!}{
\begin{NiceTabular}{cl|ccc|ccc|c|cc}
\CodeBefore
    \rowcolor{Gray}{8,13}
    \cellcolor{white}{8-1,13-1}
\Body
\toprule
\rowcolor{white}
& & \multicolumn{3}{c}{\textbf{Label Matching}} & \multicolumn{3}{c}{\textbf{Modality Matching}} & \textbf{Text} & \multicolumn{2}{c}{\textbf{Time Series}} \\ 
& Retriever & P@1 ($\uparrow$) & P@5 ($\uparrow$)& MRR ($\uparrow$) & P@1 ($\uparrow$) & P@5 ($\uparrow$) & MRR ($\uparrow$) & ROUGE ($\uparrow$) & MAE ($\downarrow$)& MSE ($\downarrow$) \\
\midrule
& Random  &42.61&47.50&0.583&0.00&0.00&0.00&0.416&0.874&1.653 \\
\midrule
\multirow{5}{*}{\rotatebox[origin=c]{90}{\textbf{TS-to-Text}}} 
& \textit{w/} Time-MoE &46.46&43.98&0.612&1.79&5.93&0.052&0.482&0.837&1.607 \\
& \textit{w/} Timer-XL &36.34&38.16&0.543&4.29&12.61&0.090&0.482&0.793&1.493 \\
& \textit{w/} TS2Vec &50.47&48.72&0.651&4.37&14.57&0.112&0.503&0.784&1.462 \\

& \textit{w/} Moment &55.73&53.18&0.691&7.78&21.68&0.154&0.515&0.747&1.415 \\
& \name &\textbf{90.08}&\textbf{77.60}&\textbf{0.940}&\textbf{44.10}&\textbf{70.24}&\textbf{0.560}&\textbf{0.717}&\textbf{0.403}&\textbf{0.771} \\

\midrule
\multirow{5}{*}{\rotatebox[origin=c]{90}{\textbf{Text-to-TS}}} 
& \textit{w/} Time-MoE &57.08&52.22&0.656&0.75&2.89&0.031&0.460&0.857&1.578 \\
& \textit{w/} Timer-XL &63.91&58.71&0.731&2.94&9.47&0.073&0.463&0.821&1.568 \\
& \textit{w/} TS2Vec &60.28&56.41&0.706&7.42&23.70&0.184&0.471&0.806&1.490 \\
& \textit{w/} Moment &64.67&59.53&0.740&5.83&18.15&0.133&0.488&0.778&1.467\\
& \name &\textbf{89.63}&\textbf{78.39}&\textbf{0.938}&\textbf{43.72}&\textbf{69.84}&\textbf{0.557}&\textbf{0.713}&\textbf{0.411}&\textbf{0.793} \\
\bottomrule
\end{NiceTabular}}\vspace{-0.5cm}
\end{table}

\subsection{Timeseries-to-Timeseries Retrieval}\label{sec:ts2ts}
To further assess the effectiveness of \name, we conduct a TS-to-TS retrieval task where each query is matched against all other time series to identify the most semantically similar ones. The evaluation is performed using the label matching metrics (Sec.~\ref{sec:retrieval}), including Precision@1, Precision@5, and Mean Reciprocal Rank (MRR), alongside query time as a proxy for computational efficiency.

\textbf{Baseline Setup.} We compare \name, against several representative time series retrieval methods. Euclidean Distance (ED) serves as a simple statistical baseline based on mean-pooled raw time series. Dynamic Time Warping (DTW), a classic elastic matching method, evaluates similarity by aligning sequences with potential shifts. SAX-VSM~\cite{senin2013sax} leverages symbolic aggregation and vector space modeling to convert time series into symbolic representations for efficient textual retrieval. CTSR~\cite{yeh2023efficient} is a learnable baseline that uses contextual metadata to enhance retrieval.

\begin{wraptable}{r}{0.4\linewidth}
\centering \vspace{-0.4cm}
\caption{TS-to-TS Retrieval performance comparison. Evaluation is conducted over 1000 randomly sampled weather time-series queries.}
\label{tab:ts2ts_retrieval}
\resizebox{\linewidth}{!}{
\begin{tabular}{lcccc}
\toprule
\textbf{Method} & \textbf{P@1} & \textbf{P@5} & \textbf{MRR} & \textbf{Time (s)} \\
\midrule
ED  & 0.548 & 0.762 & 0.644 & 0.083 \\
DTW  & 0.380 & 0.770 & 0.543 & 2273.93 \\
SAX-VSM     & 0.551 & 0.769 & 0.649 & 0.343 \\
CTSR & 0.682 & 0.893 & 0.802 &  0.057\\ 
\name& \textbf{0.900} & \textbf{0.986} & \textbf{0.938} & \textbf{0.045} \\
\bottomrule
\end{tabular}}\vspace{-0.4cm}
\end{wraptable}
\textbf{Analysis. } As shown in Table~\ref{tab:ts2ts_retrieval}, \name substantially outperforms all baselines across accuracy metrics while maintaining the lowest retrieval latency. Notably, despite the design simplicity of SAX-VSM and its moderate performance gains over raw ED, it fails to capture deep temporal or semantic patterns. CTSR, while benefiting from structured cues, struggles to generalize as effectively in purely time-series scenarios. The results suggest that \name, when equipped with task-driven objectives and textual alignment-aware training, provides not only superior retrieval quality but also enables scalable and efficient retrieval pipelines. A detailed case study on TS-to-TS retrieval is presented in Appendix~\ref{app:ts2ts}, illustrating how \name's structured aggregation across all channels effectively captures global semantics and highlights the most semantically dominant channels contributing to retrieval relevance.

\subsection{Retrieval-augmented Time Series Forecasting}\label{sec:rag}
\begin{wraptable}{r}{0.55\textwidth}
\centering\vspace{-0.5cm}
\caption{Forecasting performance on Weather dataset for next 24 steps under different retrieval-augmented generation settings.}\label{tab:RAGforecasting}
\resizebox{\linewidth}{!}{
\begin{NiceTabular}{l|cccc|cccc}
\toprule
& \multicolumn{2}{c}{\textbf{Timer-XL}} & \multicolumn{2}{c}{\textbf{Time-MoE}} & \multicolumn{2}{c}{\textbf{Moment}} & \multicolumn{2}{c}{\texttt{\textbf{\name}}} \\
\cmidrule{2-9}
 Setting& MAE & MSE & MAE & MSE & MAE & MSE & MAE & MSE \\
\midrule
\textit{w/o} RAG& 0.729 & 1.055 & 0.635 & 0.903 & 0.645 & 0.816 & 0.576 & 0.718 \\
\textit{w/} TS-only  & 0.720 & 1.009 & 0.621 & 0.801 & 0.628 & 0.797 & 0.556 & 0.698 \\
\textit{w/} TS+Text  & 0.712 & 0.984 & 0.611 & 0.787 & 0.631 & 0.801 & 0.555 & 0.696 \\
\bottomrule
\end{NiceTabular}}
\vspace{-0.3cm}
\end{wraptable}
\textbf{Setup.} We use \name to retrieve the most relevant timeseries–text pairs from the curated corpus based on time-series embedding similarity, which is then passed through trainable linear layers to produce a soft prompt. For \textit{TS-only} setting, the prompt is derived solely from the retrieved raw time series, denoted as $h_\text{ts}$; for \textit{TS+Text}, we concatenate $h_\text{ts}$ and semantic embedding $\mathbf{z}_{\mathrm{cxt}}$ from the retrieved text to form the prompt. This soft prompt is then prepended to the query for the downstream forecasting layer, without fine-tuning the pre-trained model weights. We refer to Appendix~\ref{app:RAG} for implementation details. We test two architecture families: (1) decoder-only models, including Timer-XL and Time-MoE, where the prompt is prepended at every autoregressive generation step, and (2) encoder-only models, Moment and \name, where the prompt is prepended to the encoder's hidden states and followed by a trainable forecasting head. In all settings, only the linear projection layers for prompt generation and the forecasting head (for encoder-only) are trained. 

\textbf{Results.} Table~\ref{tab:RAGforecasting} presents the forecasting results across decoder-only and encoder-only models under different RAG settings, augmented by top-$R$ retrieved instances. We refer to Figure~\ref{fig:four_ablation} (d) for ablation on $R$. The results reveal that retrieval augmentation consistently improves forecasting performance across all models, and the \textit{TS+Text} setting leads to the most significant gains for decoder-only models like Timer-XL and Time-MoE. Notably, \name shows marginal improvement when moving from \textit{TS-only} to \textit{TS+Text} retrieval, which can be attributed to that its multimodal embedding space is already aligned with textual descriptions. This alignment reduces the dependency on additional textual signals and justifies \name’s design as a lightweight, general-purpose retriever for RAG pipelines. Moreover, these results indicate decoder-only models are more sensitive to the richness of retrieved modalities, whereas encoder-only models exhibit more stable and better capacity for internalizing and utilizing structured representations. While our RAG design adopts a simple embedding concatenation strategy, it primarily validates the general utility of retrieved content across different model families. We leave optimizing augmentation architectures for future work.

\subsection{Standalone Time Series Encoder}\label{sec:encoder}

\textbf{Setup.} To evaluate \name as a standalone encoder, we conduct experiments on forecasting and classification tasks. We compare \name against full-shot models all trained from scratch, and time series foundation models. All foundation models are evaluated in a zero-shot setting, except for Moment and \name, which are fine-tuned on the forecasting head following the official protocol for forecasting. For classification, we evaluate on our curated weather dataset and fine-tune all foundation models in the same setting to ensure a fair comparison (detailed in Appendix~\ref{app:standalone_encoder}).

\begin{wraptable}{r}{0.3\textwidth} \vspace{-0.5cm}
\centering
\setlength{\tabcolsep}{1pt}
\caption{Weather Event Classification Results.}\label{tab:classification}\vspace{-0.3cm}
\resizebox{\linewidth}{!}{
\begin{tabular}{lcc}
\toprule
\textbf{Model} & \textbf{Accuracy}  & \textbf{F1} \\
\midrule
\multicolumn{3}{l}{\textit{\textbf{Train-from-scratch Model}}} \\
DLinear             & 82.37 & 65.78 \\
iTransformer        & 84.99 & 68.29 \\
PatchTST            & 84.78 & 69.13 \\
TimesNet            & 86.09 & 68.97 \\
TimeMixer           & 84.78 & 68.65 \\
FSCA                & 85.62      & 69.41       \\
\midrule
\multicolumn{3}{l}{\textit{\textbf{Finetune a Pre-trained Model}}} \\
Time-MoE$_\text{large}$& 59.09     & 19.74     \\
Moment$_\text{base}$& 65.43     & 28.29     \\
Timer-XL               & 72.38   & 33.45    \\
Chronos$_\text{tiny}$& 74.79     & 40.21     \\
\rowcolor{Gray}
\name \small{\textit{w/o} RAG} & 85.20 &  69.98\\
\rowcolor{Gray}
\name \small{\textit{w/} RAG}  & \textbf{89.76} & \textbf{72.36} \\
\bottomrule
\end{tabular}}
\vspace{-0.5cm}
\end{wraptable}
\textbf{Results.} As shown in Table~\ref{tab:forecasting}, \name outperforms baselines across different datasets and showcases capability on longer forecasting horizons ($H$), whereas the performance of baselines exhibits considerable variation. This observation justifies the cross-modal design behind \name, which equips the model with stronger semantic grounding and context-aware forecasting. In the event-type classification task (as shown in Table~\ref{tab:classification}), we observe that fine-tuned foundation models underperform traditional train-from-scratch baselines, suggesting that their embeddings may be overgeneralized and poorly adapted to domain-specific classification signals. In contrast, \name achieves significantly higher accuracy and F1 without RAG, and benefits further from the retrieval-augmented setting. This demonstrates \texttt{TRACE}’s ability to retain discriminative structure while maintaining broad semantic alignment, which is essential for robust downstream deployment. Full results of other foundation model variants are in Appendix~\ref{app:classification_results}.

\begin{table}[t]
\vspace{-0.3cm}
\centering
\caption{Forecasting results (MAE and MSE) of full-shot models and time series foundation models on multi-variate (\textbf{M}) and univariate (\textbf{U}) datasets. \textbf{\color{red}Red}: the best, \underline{\color{blue}Blue}: the 2nd best.}\label{tab:forecasting}
\setlength{\tabcolsep}{4pt} 
\resizebox{\linewidth}{!}{
\begin{NiceTabular}{ll*{4}{|cccc}|c}
\CodeBefore
\rowcolor{Gray}{16}
\Body
\toprule
& \multirow{3}{*}{Model} 
& \multicolumn{4}{c}{\textbf{Weather (M)}} 
& \multicolumn{4}{c}{\textbf{Health (U)}} 
& \multicolumn{4}{c}{\textbf{Energy (U)}} 
& \multicolumn{4}{c}{\textbf{Environment (U)}} 
& \multirow{3}{*}{\makecell{\# \\\textbf{1}$^{\text{st}}$}} \\
\cmidrule{3-18}
&& \multicolumn{2}{c}{H = 7} & \multicolumn{2}{c}{H = 24} 
   & \multicolumn{2}{c}{H = 12} & \multicolumn{2}{c}{H = 48} 
   & \multicolumn{2}{c}{H = 12} & \multicolumn{2}{c}{H = 48} 
   & \multicolumn{2}{c}{H = 48} & \multicolumn{2}{c}{H = 336} 
   & \\
&& MAE & MSE & MAE & MSE 
   & MAE & MSE & MAE & MSE 
   & MAE & MSE & MAE & MSE 
   & MAE & MSE & MAE & MSE & \\
\toprule
\multirow{6}{*}{\rotatebox[origin=c]{90}{\textbf{Zero-shot}}}
&Chronos & 0.560 & 0.937 & 0.646 & 1.094 & 0.650 & 1.106 & 0.987 & 2.019 & 0.263 & 0.148 & 0.554 & 0.553 & 0.536 & 0.612 & 0.583 & 0.671 &0 \\
&Time-MoE & 0.579 & 0.803 & 0.635 & 0.903 & \textcolor{blue}{\underline{0.604}} & 0.981 & \textcolor{blue}{\underline{0.832}} & 1.697 & \textcolor{red}{\textbf{0.205}} & \textcolor{red}{\textbf{0.089}} & \textcolor{blue}{\underline{0.451}} & \textcolor{blue}{\underline{0.396}} & 0.562 & 0.508 & 0.836 & 0.969 & \textcolor{blue}{\underline{2}}\\
&TimesFM & 0.550 & 0.859 & 0.640 & 1.034 & 0.610 & \textcolor{blue}{\underline{0.913}} & 0.865 & \textcolor{blue}{\underline{1.685}} & 0.248 & 0.137 & 0.499 & 0.482 & 0.503 & 0.532 & 0.531 & 0.569 &0\\
&Timer-XL & 0.645 & 0.912 & 0.729 & 1.055 & 0.741 & 1.235 & 0.988 & 1.892 & 0.236 & 0.118 & 0.460 & 0.424 & 0.549 & 0.564 & 0.565 & 0.574 &0 \\
&Moirai & 0.593 & 1.001 & 0.675 & 1.135 & 0.976 & 3.029 & 1.569 & 8.125 & 0.318 & 0.273 & 0.692 & 1.415 & 0.935 & 12.428 & 2.237 & 25.011 &0 \\
&Moment&0.572 &0.732 &0.645 &0.816& 0.988 & 1.824 & 0.997& 1.902 & 0.471 & 0.411 & 0.542 & 0.542 & \textcolor{red}{\textbf{0.449}} & \textcolor{red}{\textbf{0.375}} & 0.554 & 0.502 & \textcolor{blue}{\underline{2}}\\
\midrule
\multirow{7}{*}{\rotatebox[origin=c]{90}{\textbf{Full-shot}}}
&DLinear & 0.593 & 0.778 & 0.691 & 0.884 & 1.178 & 2.421 & 1.132 & 2.256 & 0.410 & 0.273 & 0.546 & 0.512 & 0.561 & 0.515 & 0.581 & 0.534 &0 \\
&iTransformer & 0.518 & 0.707 & 0.591 & 0.814 & 0.676 & 1.072 & 0.911 & 1.747 & 0.267 & 0.124 & 0.487 & 0.399 & 0.486 & 0.425 & 0.511 & 0.458 &0 \\
&PatchTST & 0.529 & 0.723 & 0.599 & 0.826 & 0.656 & 1.034 & 0.902 & 1.708 & 0.263 & 0.121 & 0.489 & 0.407 & 0.493 &0.462 & 0.525 & 0.511 &0 \\
&TimesNet & \textcolor{blue}{\underline{0.497}} & 0.654 & \textcolor{blue}{\underline{0.581}} & 0.786 & 0.820 & 1.376 & 0.969 & 1.903 & 0.270 & 0.127 & 0.496 & 0.398 & 0.520 & 0.486 & \textcolor{blue}{\underline{0.489}} & \textcolor{blue}{\underline{0.430}}&0 \\
&TimeMixer & 0.501 & 0.667 & 0.585 & 0.787 & 1.091 & 2.215 & 1.126 & 2.250 & 0.376 & 0.246 & 0.538 & 0.491 & 0.558 & 0.553 & 0.559 & 0.568 &0 \\
&FSCA &\textcolor{red}{\textbf{0.496}} &\textcolor{blue}{\underline{0.642}} &0.780  &\textcolor{blue}{\underline{0.762}}  & 0.756 & 1.240 & 0.969 & 1.904 & 0.278 & 0.136 & 0.520 & 0.466 & 0.497 & 0.462 & 0.511 & 0.496 &1 \\
&\textbf{\name}& 0.501 & \textcolor{red}{\textbf{0.623}} & \textcolor{red}{\textbf{0.576}}& \textcolor{red}{\textbf{0.718}} & \textcolor{red}{\textbf{0.547}} & \textcolor{red}{\textbf{0.768}} & \textcolor{red}{\textbf{0.827}} & \textcolor{red}{\textbf{1.435}} & \textcolor{blue}{\underline{0.230}}& \textcolor{blue}{\underline{0.113}} &\textcolor{red}{\textbf{0.448}}  &\textcolor{red}{\textbf{0.389}}  & \textcolor{blue}{\underline{0.455}} & \textcolor{blue}{\underline{0.403}} & \textcolor{red}{\textbf{0.475}}& \textcolor{red}{\textbf{0.413}} &\textcolor{red}{\textbf{11}}\\
\bottomrule
\end{NiceTabular}}\vspace{-0.5cm}
\end{table}

\section{Ablation Studies}
\label{section:case_studies}
\begin{figure}[htbp]\vspace{-0.3cm}
    \centering
    \includegraphics[width=\linewidth]{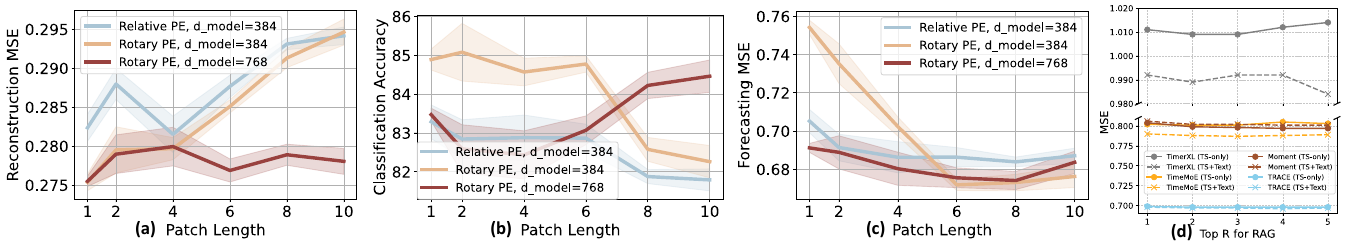}
    \caption{Ablation studies on patch length, positional embedding, and hidden dimension for (a) Reconstruction MSE, (b) Classification Accuracy (\%), and (c) Average Forecasting MSE. (d) shows ablation studies on the number of retrieved instances ($R$) in the RAG pipeline. }
    \label{fig:four_ablation}\vspace{-0.5cm}
\end{figure}

\textbf{Hyper-parameter Sensitivity.} Figure~\ref{fig:four_ablation} presents a comprehensive ablation study investigating the effects of patch length $P$, positional embedding (PE) types, and the number of retrieved instances $R$ used in our RAG setup. Rotary PE consistently outperforms Relative PE by achieving lower reconstruction and forecasting MSEs as well as higher classification accuracy, particularly when using a smaller model size ($d=384$). Notably, increasing the model size to $d=768$ does not yield significant improvements, especially for downstream forecasting and classification tasks, suggesting that careful architectural design and PE choice may matter more than simply scaling parameters. Across tasks, mid-range patch lengths (\eg $P=6$) offer the best trade-off between local and global temporal resolution. In Figure~\ref{fig:four_ablation} (d), we observe that time series foundation models are relatively robust to the choice of $R$, and models augmented with aligned text generally outperform their TS-only counterparts, highlighting the benefit of cross-modal retrieval in improving forecasting performance.

\begin{wrapfigure}{r}{0.45\textwidth}\vspace{-0.3cm}
\begin{minipage}{\linewidth}
  \centering
  \captionof{table}{Ablation study on attention variants in pre-training architecture.}
  \label{tab:ablation_attention}
  \resizebox{\linewidth}{!}{
  \begin{tabular}{l cc}
    \toprule
    & Avg. MSE & Acc. (\%) \\
    \midrule
    \name &\textbf{0.670$\pm$0.013}  &\textbf{85.20$\pm$0.13}\\
    \textit{w/o} \texttt{CIT} &0.713$\pm$0.016 &85.04$\pm$0.26\\
    \texttt{CbA} $\Rightarrow$ Full Attn &0.705$\pm$0.013 & 84.18$\pm$0.11\\
    \texttt{CbA} $\Rightarrow$ Causal Attn &0.682$\pm$0.015&83.72$\pm$0.13 \\
    \bottomrule
  \vspace{0.2cm}
  \end{tabular}}
  \includegraphics[width=\linewidth]{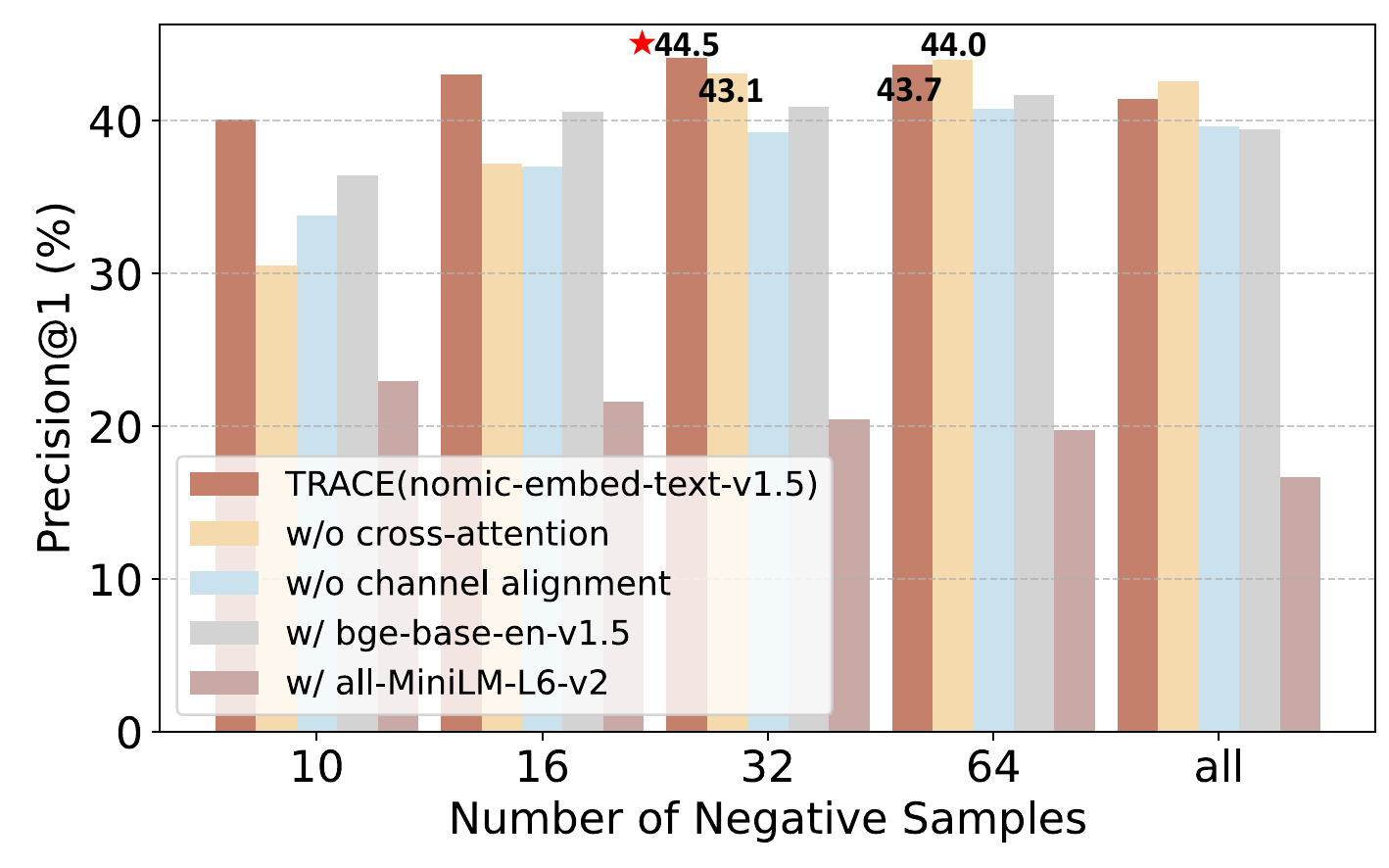}
  \captionof{figure}{Retrieval performance under varying numbers of negative samples. The best is indicated by {\color{red}$\star$}.}
  \label{fig:ablation_N}
\end{minipage} \vspace{-0.5cm}
\end{wrapfigure}

\textbf{Attention Variants.} Table~\ref{tab:ablation_attention} assess the impact of key architectural choices in \name, including channel identity token (\texttt{CIT}) and different attention mechanisms. Removing \texttt{CIT} results in a notable increase in average MSE, indicating its importance for capturing fine-grained temporal dependencies. We also replace the channel-biased attention (\texttt{CbA}) with two alternatives: full attention, similar to a multivariate variant of Moment~\cite{goswami2024moment}, and causal attention, analogous to decoder-only designs like Timer-XL~\cite{liu2024timer}, Both alternatives yield degraded performance. These results highlight the effectiveness of the architectural design in \name, particularly the synergy between \texttt{CIT} and \texttt{CbA} in achieving outstanding performance. We refer to Appendix~\ref{app:efficiency} for runtime and efficiency evaluation.

\textbf{Cross-Attention and Hard Negative Sampling.} Figure~\ref{fig:ablation_N} presents an ablation study on key components in \name for retrieval precision under different numbers of negative samples ($K$). “all” indicates using the entire batch (excluding the paired counterpart) as negatives. The default model, using \texttt{nomic} text encoder~\cite{nussbaum2024nomic}, consistently achieves the highest performance, especially when $K$ is small, highlighting its efficacy in low-computation settings. Removing the final cross-attention module between time series and text leads to notable performance degradation under small $K$, suggesting that cross-modal fusion becomes especially crucial when fewer negatives are available. Similarly, eliminating channel-level alignment yields a consistent drop, confirming the strength of the proposed dual-level contrastive mechanism. Substituting \texttt{nomic} with weaker text encoders like \texttt{bge} or \texttt{MiniLM} results in worse performance, implying that high-quality embeddings are necessary for discriminating harder negatives. Overall, these trends support the effectiveness of our hard negative mining strategy and emphasize the importance of dual-level alignment in retrieval performance. We provide empirical case studies in Appendix~\ref{app:case_study}.

\section{Conclusion and Future Work}
\label{sec:conclusion}
We introduce \name, a multimodal framework that aligns time series with textual descriptions at both channel and sample levels. Extensive experiments demonstrate that \name outperforms strong baselines across retrieval, standalone encoding, retrieval-augmented settings, and generalizes well across model families. One limitation is that \name relies on supervised textual alignment, which may not be readily available in all domains. In future work, we plan to extend \name to support weakly supervised and semi-supervised settings, where textual context is partially missing or noisy. Another promising direction is integrating domain adaptation techniques to improve generalization across unseen domains and sensor modalities (\eg image, video). Moreover, exploring autoregressive generation conditioned on retrieved time series–text pairs may further enhance understanding tasks in temporal modeling. We refer to Appendix~\ref{app:discussion} for a detailed discussion on social impact.

\section*{Acknowledgments and Disclosure of Funding} 
This research was supported in part by the National Science Foundation (NSF) Division of Computer and Network Systems (CNS-2431504), Army Research Office contract W911NF-23-1-0088, the Yale AI Engineering Research Grant from the Yale Office of the Provost, the LEAP-U Sponsored Research from Samsung Research America, the Roberts Innovation Award 2025, and the AWS Research Awards. We would also like to thank the anonymous reviewers for their insightful and constructive feedback.

\newpage
\bibliographystyle{unsrt}
\bibliography{reference}



\newpage
\appendix
\begin{center}
    \LARGE \textbf{Appendix}
\end{center}

\section{Notations}\label{app:notation}
The main notations used throughout this paper are summarized in Table~\ref{tab:notation}.
\begin{table}[th]
    \centering
    \caption{Summary of the notations used in this paper.}
    \label{tab:notation}
    \begin{tabular}{c|l}
    \toprule
    \textbf{Notation} & \textbf{Description} \\
    \midrule
    $\mathbf{X} \in \mathbb{R}^{C \times T}$ & Input multivariate time series with $C$ channels and $T$ time steps \\
    $P$ & Patch length for time series tokenization \\
    $\hat{T}$ & Number of temporal patches per channel ($\lfloor T / P \rfloor$) \\
    $C$ & Number of channels (variables) \\
    $T$ & Number of time steps in the original sequence \\
    $H$ & Forecasting horizon \\
    $d$ & Embedding dimension \\
    $L$ & Length of the flattened token sequence \\
    $X_i^{\text{patch}}$ & Embedding of the $i$-th patch token \\
    $M\in\{0,1\}^{L\times L}$ & Biased attention mask for the flatten token sequence \\
    $\mathbf{H}^{\text{out}} \in \mathbb{R}^{L \times d}$ & Output token embeddings from the Transformer encoder \\
    $\mathbf{h}_{\texttt{[CLS]}} \in \mathbb{R}^d$ & Embedding of the global [CLS] token \\
    $\mathbf{H}_{\texttt{[CIT]}} \in \mathbb{R}^{C \times d}$ & Embeddings of Channel Identity Tokens (CITs) \\
    $\tau_{\text{cxt}}$ & Sample-level textual context associated with a time series \\
    $\tau_c$ & Channel-level textual description for the $c$-th variable \\
    $\mathbf{z}_{\text{cxt}} \in \mathbb{R}^d$ & Semantic embedding of the sample-level text \\
    $\mathbf{z}_c \in \mathbb{R}^d$ & Semantic embedding of the channel-level text \\
    $\mathcal{N}_{\text{cxt}}, \mathcal{N}_{\text{ch}}$ & Sample-level and channel-level hard negative candidate sets\\
    \bottomrule
    \end{tabular}
\end{table}

\section{Dataset Curation}\label{app:our_dataset}
We curate a new multimodal time series dataset in the weather domain by extending MTBench \cite{chen2025mtbenchmultimodaltimeseries}. It is built from two primary sources:
\begin{itemize}
    \item \textbf{Event reports} from the \textit{NOAA Storm Events Database}\cite{stormevents}, which contains detailed narratives of severe weather occurrences across the U.S.
    \item \textbf{Weather Time Series (TS) data} from the \textit{NOAA Global Historical Climatology Network - Hourly (GHCN-h)}\cite{ghcnhourly2023}, covering multiple meteorological variables.
\end{itemize}
When applying \name to our curated dataset, the sample-level context is event report, while the channel-level description is synthetically generated by LLMs.
\subsection{Station and Event Selection}
We begin by selecting over 100 U.S. locations frequently affected by severe weather events and associated with long narrative reports. This yields approximately 5,000 event entries. For each event location, we identify nearby GHCN-h weather stations and extract multivariate TS data anchored at the start time of each event.

\subsection{Time Series Sampling}

\begin{figure}[h]
    \centering
    \includegraphics[width=1\linewidth]{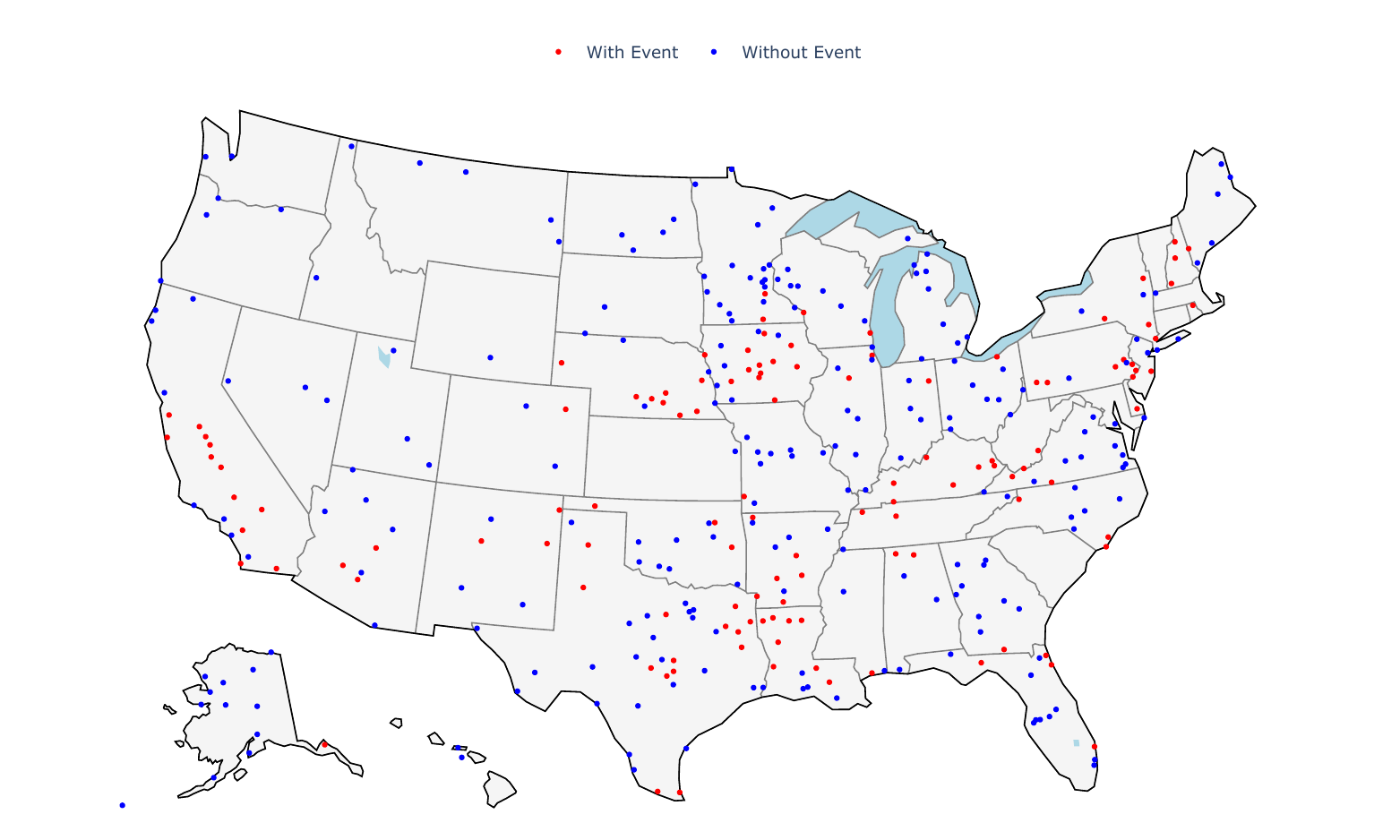}
    \caption{The red points are locations with event reports and the blue points are locations without event reports}
    \label{fig:event_map}
\end{figure}

Each event is treated as an anchor point to extract TS data at three resolutions:
\begin{itemize}
    \item Hourly for 7 days
    \item Every 4 hours for 28 days
    \item Daily for 180 days
\end{itemize}
This results in approximately 15,000 TS samples from event-associated windows. To balance the dataset, we sample an additional 30,000 TS sequences from the same stations at random non-event times, ensuring no overlapping event narratives. To enhance weather diversity, we also sample 30,000 TS sequences from geographically distant stations without any event association, using randomly selected anchor times. See Figure \ref{fig:event_map}. All time series instances contain seven channels: temperature, humidity, wind\_u, wind\_v, visibility, precipitation, and sky code. The curated weather dataset contains a total of 74,337 time series instances, and the lengths have a mean of 169.25 and a median of 168.0.
\begin{Mycolorbox}[Example: An Event Report of Debris Flow]{}
    "\texttt{event type}": "Debris Flow", \\
    "\texttt{}{state}": "CALIFORNIA", \\
    "\texttt{cz name}": "TULARE", \\
    "\texttt{begin datetime}": "2021-12-14 13:14:00", \\
    "\texttt{end datetime}": "2021-12-14 16:14:00", \\
    "\texttt{narrative}": "A strong low pressure system dropped southeast out of the Gulf of Alaska on December 12 and intensified off the Pacific Northwest coast on December 13 pulling up some deep moisture which was pushed into central California during the afternoon.  The precipitation intensified during the evening of December 13 through the morning of December 14 as the low carved out a deep upper trough which pushed across California during the afternoon of December 14.  This system produced 2 to 4 inches of liquid precipitation over the Sierra Nevada from Sequoia National Park northward and 1 to 3 inches of liquid precipitation south of Sequoia Park.  The precipitation fell mainly in the form of snow above 5500 feet and several high elevation SNOTELs estimated 2 to 4 feet of new snowfall.  The snow level lowered to as low as 1500 feet during the evening of December 14 as the cooler airmass behind the system pushed into central California.  Much of the San Joaquin Valley picked up between 1 to 2 inches of rainfall while the Kern County Mountains picked up between 0.75 and 1.5 inches of liquid precipitation.  The Kern County Desert areas only picked up between a quarter and a half inch of rain at most locations due to rain shadowing.  The storm produced widespread minor nuisance flooding in the San Joaquin Valley and Sierra foothills with a few rock slides noticed.  Several roads were closed as a precaution and chain restrictions were implemented on some roads in the Sierra Nevada.  The storm also produced strong winds over the West Side Hills as well as in the Grapevine and Tehachapi areas in Kern County.  Several stations in these areas measured wind gusts exceeding 50 mph with a few locations near the Grapevine measuring brief gusts exceeding 70 mph. California Highway Patrol reported mud, rock and dirt covering most of North Plano St. near Lynch Dr.",
\end{Mycolorbox}

\subsection{Synthetic Description Generation}

We use ChatGPT to generate channel-level textual descriptions for selected TS samples, where all TS samples linked to event reports are included. We also randomly select 50\% of TS samples from both the non-event windows at event-associated stations and the non-event-associated stations to generate channel-level descriptions for event-label balance. The generated descriptions follow the style of TimeCap\cite{lee2025timecaplearningcontextualizeaugment}, but each is additionally annotated with one or more keywords selected from the set as auxiliary information: \{\text{Clear}, \text{Cloudy}, \text{Rainy}, \text{Snowy}, \text{Windy}, \text{Foggy}, \text{Hot}, \text{Cold}, \text{Humid}, \text{Stormy}\}.
We use a consistent meta-prompt to elicit both descriptive and label-aligned outputs. A full example of the meta-prompt and a generated description is provided in \ref{sec:prompt and description}.
\subsection{Prompt for Weather Description Generation and an example synthetic description}
\label{sec:prompt and description}
\begin{Mycolorbox}[Weather Summary Prompt]{}
You are a daily weather reporter, asked to summarize the past seven days of hourly weather (or the past 28 days of 4-hourly weather, or the past 6 months of daily weather, depending on the selected mode).

It will be multichannel with \texttt{temperature}, \texttt{precipitation}, \texttt{relative\_humidity}, \texttt{visibility}, \texttt{wind\_u}, and \texttt{wind\_v} aspects. Summarize these channels and label the overall weather with one or more keywords from the set:

\{Clear, Cloudy, Rainy, Snowy, Windy, Foggy, Hot, Cold, Humid, Stormy\}.

You are \textbf{not} expected to report each time point individually. Instead, analyze the entire period as a whole. Additionally, for \texttt{temperature}, \texttt{precipitation}, and \texttt{relative\_humidity}, identify any noticeable trends, potential periodicities (\eg daily or weekly patterns), overall volatility, and any clear outliers that stand out. You do not need to analyze other channels for these advanced statistics.

The input includes:

\begin{itemize}
    \item Location
    \item Date
    \item Temperature time series
    \item Precipitation time series
    \item Relative humidity time series
    \item Visibility time series
    \item Wind\_u time series
    \item Wind\_v time series
    \item Sky cover codes
\end{itemize}

Sky cover codes are interpreted as follows:

\begin{center}
\begin{tabular}{lll}
\textbf{Code} & \textbf{Meaning} & \textbf{Sky Fraction Covered} \\
00 & CLR (Clear) & 0/8 or 0\% \\
01 & FEW & 1/8 (~12\%) \\
02 & FEW & 2/8 - 3/8 (25\%-37\%) \\
03 & SCT (Scattered) & 4/8 (~50\%) \\
04 & SCT & 5/8 (~62\%) \\
05 & BKN (Broken) & 6/8 (~75\%) \\
06 & BKN & 7/8 - 8/8 (87\%-100\%) \\
07 & BKN & $\sim$9/10 \\
08 & OVC (Overcast) & 10/10 (100\%) \\
09 & VV (Vertical Visibility) & Sky obscured \\
10 & X (Unknown) & Partially obscured \\
\end{tabular}
\end{center}

\textbf{Please summarize the data using the following format:}

\begin{itemize}
    \item \textbf{Date:} \{sentence of date\}
    \item \textbf{Location:} \{sentence of location\}
    \item \textbf{Temperature:} \{sentence of temperature\}
    \item \textbf{Precipitation:} \{sentence of precipitation\}
    \item \textbf{Relative Humidity:} \{sentence of relative humidity\}
    \item \textbf{Visibility:} \{sentence of visibility\}
    \item \textbf{Wind\_V:} \{sentence of wind\_v\}
    \item \textbf{Wind\_U:} \{sentence of wind\_u\}
    \item \textbf{Sky Cover:} \{sentence of sky cover\}
    \item \textbf{Keywords:} \{list of keywords from label set\}
\end{itemize}

No additional explanation or commentary should be included in the output.
\end{Mycolorbox}

\begin{Mycolorbox}[Example: Generated Weather Summary]{}
\textbf{Date:} The past 28 days from January 30, 2021, to February 26, 2021.

\textbf{Location:} The weather data is from Pike, Kentucky.

\textbf{Temperature:} The temperature ranged from a low of -10.07\,\textdegree C to a high of 20.25\,\textdegree C, with noticeable fluctuations and a general upward trend towards the end of the period, indicating warming conditions.

\textbf{Precipitation:} There was no recorded precipitation throughout the 28 days, indicating dry weather.

\textbf{Relative Humidity:} Relative humidity varied significantly, peaking at 100\% on multiple occasions, with a general trend of higher humidity levels during the earlier part of the period and lower levels towards the end.

\textbf{Visibility:} Visibility remained consistently high at 14.58\,km throughout the reporting period.

\textbf{Wind\_V:} Wind velocity showed variability, with occasional gusts and a general trend of calm conditions.

\textbf{Wind\_U:} Wind direction fluctuated, with both positive and negative values indicating changes in wind patterns.

\textbf{Sky Cover:} The sky was consistently clear with no significant cloud cover reported.

\textbf{Keywords:} [Clear, Cold, Humid]
\end{Mycolorbox}

\subsection{Dataset Details}\label{app:dataset_stat}
\begin{table}[ht]
\caption{Dataset size for each task.}\label{tab:data_lengths}
\centering
\begin{tabular}{lrrrr}
\toprule
\textbf{Dataset Type} & \textbf{Train} & \textbf{Test} & \textbf{Val} & \textbf{Total} \\
\midrule
\multicolumn{5}{l}{\textbf{\textit{Newly Curated Weather Dataset}}} \\
{ }Forecasting (H=7) & 6,690 & 957 & 1,914 & 9,561 \\
{ }Pretraining \& Classification& 45,339 & 6,484 & 12,953 & 64,776\\
\midrule
\midrule
\multicolumn{5}{l}{\textbf{\textit{Public Dataset from TimeMMD~\cite{liu2024time}}}} \\
Health (H=12) & 929 & 266 & 129 &1,324\\
Energy (H=12) &992 & 284&138&1,414 \\ 
Environment (H=48) &7,628&2,173&1,064&10,865 \\
\bottomrule
\end{tabular}
\end{table}
Our curated weather dataset contains a total of 74,337 time series instances. We allocate 9,561 of these exclusively for the forecasting task, ensuring this subset is disjoint from the pretraining and classification data to avoid any potential label leakage or information overlap. The classification task is formulated as multi-class event prediction, where each time series instance is annotated by the NOAA System with a corresponding weather event type from nine common severe weather events, and one special category for non-events. The event labels are as follows:
\textit{Lightning (0), Debris Flow (1), Flash Flood (2), Heavy Rain (3), Tornado (4), Funnel Cloud (5), Hail (6), Flood (7), Thunderstorm Wind (8)}.
Instances that do not correspond to any specific event are labeled as \texttt{None}. This setup ensures the model learns to distinguish between distinct event types while being robust to trivial (non-event) data.  We follow the original split to create train/test/val set for TimeMMD forecasting tasks~\cite{liu2024time}. 

\section{Alignment Objective: Full Formulation}\label{app:method_details}

To fully capture the structured alignment between multivariate time series and text, we employ a dual-level contrastive learning strategy, consisting of sample-level and channel-level hard negative mining.

\subsection{Hard Negative Candidate Sets}

Given a time series instance $i$ with $C$ channels, we define the following negative sets:

\paragraph{Sample-level negative sets.}
For aligning the global [CLS] embedding $\mathbf{h}_{\texttt{[CLS]}}^{(i)}$ of instance $i$ with its corresponding sample-level textual embedding $\mathbf{z}_{\mathrm{cxt}}^{(i)}$, we mine hard negatives from other samples in the batch. Specifically:
\begin{equation}
\mathcal{N}_{\mathrm{cxt}}^{(i)} = \operatorname{Top}_K \left\{ \operatorname{sim}(\mathbf{h}_{\texttt{[CLS]}}^{(i)}, \mathbf{z}_{\mathrm{cxt}}^{(j)}) \mid j \neq i \right\},
\end{equation}
and symmetrically,
\begin{equation}
\mathcal{N}_{\mathrm{cxt}}^{(i, \text{text})} = \operatorname{Top}_K \left\{ \operatorname{sim}(\mathbf{z}_{\mathrm{cxt}}^{(i)}, \mathbf{h}_{\texttt{[CLS]}}^{(j)}) \mid j \neq i \right\}.
\end{equation}

\paragraph{Channel-level negative sets.}
To align each channel-specific CIT embedding $\mathbf{h}_c^{(i)}$ with its corresponding channel-level text embedding $\mathbf{z}_c^{(i)}$, we mine two types of distractors:
\begin{itemize}
    \item \textit{Intra-instance negatives:} embeddings from other channels within the same instance, i.e., $\mathbf{z}_{c'}^{(i)}$ where $c' \neq c$;
    \item \textit{Inter-instance negatives:} same-indexed channel embeddings across different instances, i.e., $\mathbf{z}_c^{(j)}$ where $j \neq i$.
\end{itemize}
Formally, the channel-level negative set is defined as:
\begin{equation}
\mathcal{N}_{\mathrm{ch}}^{(i, c)} = \operatorname{Top}_K \left\{ \operatorname{sim}(\mathbf{h}_c^{(i)}, \mathbf{z}_{c'}^{(j)}) \mid c' \neq c \text{ or } j \neq i \right\},
\end{equation}
and similarly in the reverse direction:
\begin{equation}
\mathcal{N}_{\mathrm{ch}}^{(i, c, \text{text})} = \operatorname{Top}_K \left\{ \operatorname{sim}(\mathbf{z}_c^{(i)}, \mathbf{h}_{c'}^{(j)}) \mid c' \neq c \text{ or } j \neq i \right\}.
\end{equation}

\subsection{Contrastive Alignment Loss}

We adopt a bidirectional InfoNCE loss at both the sample and channel levels. For each alignment direction, the objective maximizes the similarity between the positive pair and minimizes similarity with hard negatives.

\paragraph{Sample-level loss.}
\begin{align}
\mathcal{L}_{\mathrm{global}}^{\text{text} \rightarrow \text{ts}} &= -\log \frac{\exp(\operatorname{sim}(\mathbf{z}_{\mathrm{cxt}}^{(i)}, \mathbf{h}_{\texttt{[CLS]}}^{(i)}) / \tau)}{
\sum\limits_{j \in \{i\} \cup \mathcal{N}_{\mathrm{cxt}}^{(i, \text{text})}} \exp(\operatorname{sim}(\mathbf{z}_{\mathrm{cxt}}^{(i)}, \mathbf{h}_{\texttt{[CLS]}}^{(j)}) / \tau)} \\
\mathcal{L}_{\mathrm{global}}^{\text{ts} \rightarrow \text{text}} &= -\log \frac{\exp(\operatorname{sim}(\mathbf{h}_{\texttt{[CLS]}}^{(i)}, \mathbf{z}_{\mathrm{cxt}}^{(i)}) / \tau)}{
\sum\limits_{j \in \{i\} \cup \mathcal{N}_{\mathrm{cxt}}^{(i)}} \exp(\operatorname{sim}(\mathbf{h}_{\texttt{[CLS]}}^{(i)}, \mathbf{z}_{\mathrm{cxt}}^{(j)}) / \tau)}
\end{align}

\paragraph{Channel-level loss.}
\begin{align}
\mathcal{L}_{\mathrm{channel}}^{\text{text} \rightarrow \text{ts}} &= \frac{1}{C} \sum_{c=1}^C -\log \frac{\exp(\operatorname{sim}(\mathbf{z}_c^{(i)}, \mathbf{h}_c^{(i)}) / \tau)}{
\sum\limits_{(j,c') \in \{(i,c)\} \cup \mathcal{N}_{\mathrm{ch}}^{(i,c,\text{text})}} \exp(\operatorname{sim}(\mathbf{z}_c^{(i)}, \mathbf{h}_{c'}^{(j)}) / \tau)} \\
\mathcal{L}_{\mathrm{channel}}^{\text{ts} \rightarrow \text{text}} &= \frac{1}{C} \sum_{c=1}^C -\log \frac{\exp(\operatorname{sim}(\mathbf{h}_c^{(i)}, \mathbf{z}_c^{(i)}) / \tau)}{
\sum\limits_{(j,c') \in \{(i,c)\} \cup \mathcal{N}_{\mathrm{ch}}^{(i,c)}} \exp(\operatorname{sim}(\mathbf{h}_c^{(i)}, \mathbf{z}_{c'}^{(j)}) / \tau)}
\end{align}

\subsection{Total Loss Objective}

The total alignment loss is the average of both sample-level and channel-level contrastive losses:
\begin{equation}
\mathcal{L}_{\text{align}} = \frac{1}{2} \left( \mathcal{L}_{\mathrm{global}}^{\text{text} \rightarrow \text{ts}} + \mathcal{L}_{\mathrm{global}}^{\text{ts} \rightarrow \text{text}} \right)
+ \lambda_{\text{ch}} \cdot \frac{1}{2} \left( \mathcal{L}_{\mathrm{channel}}^{\text{text} \rightarrow \text{ts}} + \mathcal{L}_{\mathrm{channel}}^{\text{ts} \rightarrow \text{text}} \right),
\end{equation}
where $\tau$ is the temperature hyperparameter, and $\lambda_{\text{ch}}$ is a hyperparameter, controlling the contribution of channel-level alignment. We set $\lambda_{\text{ch}}=1.0$ as default in experiments.

\section{Experiments}
\subsection{Baselines}\label{app:baseline}
\subsubsection{Full-shot Time Series Models}
\textbf{DLinear} \cite{DLinear}is a lightweight time-series forecasting model that decomposes the input into trend and seasonal components, and applies simple linear layers to each component separately. Despite its simplicity, DLinear has demonstrated strong performance on both long- and short-term forecasting tasks by effectively capturing linear temporal patterns without relying on complex neural architectures.

\textbf{PatchTST} \cite{Yuqietal-2023-PatchTST} reformulates time-series forecasting as a patch-based sequence modeling problem. It splits the input time series into non-overlapping patches and applies a Transformer encoder to model inter-patch dependencies. The design removes positional encoding and avoids decoder layers, making the model more suitable for forecasting tasks while benefiting from the global receptive field of Transformers.

\textbf{iTransformer} \cite{iTransformer} (Instance-aware Transformer) extends Transformer-based forecasting by modeling instance-wise variations. It introduces a shared backbone Transformer and an instance-specific modulation mechanism, enabling the model to better adapt to diverse temporal dynamics across different time-series samples. This design improves generalization and robustness, particularly for multivariate forecasting.

\textbf{TimesNet} \cite{wu2023timesnet} proposes a novel temporal block that captures multi-frequency patterns in time-series data using learnable convolutions in the frequency domain. By combining time and frequency-domain features, TimesNet achieves strong performance across a variety of datasets. It is particularly effective at modeling both short-term and long-term temporal dependencies.

\textbf{TimeMixer} \cite{wang2024timemixerdecomposablemultiscalemixing} employs a structured state-space-inspired architecture where time mixing and channel mixing operations alternate. It replaces self-attention with parameter-efficient mixing blocks that blend information across the temporal and feature dimensions. TimeMixer is designed for scalable forecasting and excels in low-resource regimes due to its compact architecture and efficient training.

\textbf{FSCA} \cite{hu2025context} introduces a new paradigm that aligns time series (TS) with a linguistic component in the language environments familiar to
LLMs to enable LLMs to contextualize and comprehend TS data, thereby activating their capabilities. FSCA uses a Dual-Scale Context-Alignment Graph Neural Networks (DSCA-GNNs) framework to achieve both structural and logical alignment, demonstrate good performance in few-shot and zero-shot settings.

\subsubsection{Time Series Foundation Model}
\begin{table}[ht]
\centering
\caption{Comparison of time-series foundation models.}
\label{tab:tsfm_comparison}
\resizebox{\linewidth}{!}{
\begin{NiceTabular}{l|cccccc}
\toprule
\textbf{Method} & \textbf{Chronos} & \textbf{Time-MoE} & \textbf{TimesFM} & \textbf{Moirai} & \textbf{Moment} & \textbf{Timer-XL} \\
\midrule
\textbf{Architecture} & Encoder-Decoder & Decoder-Only & Decoder-Only & Encoder-Only & Encoder-Only&Decoder-only \\
\textbf{(Max) Model Size} & 710M & 2.4B & 200M & 311M & 385M &84M\\
\textbf{Input Token} & Point & Point & Patch & Patch & Patch& Patch \\
\textbf{Max Length} & 512 & 4096 & 512 & 5000 & 512&1024 \\
\textbf{FFN} & Dense & Sparse & Dense & Dense & Dense&Dense \\
\textbf{Cross-channel} &\xmark & \xmark & \xmark & \cmark & \xmark&\cmark  \\
\bottomrule
\end{NiceTabular}
}
\end{table}
We test several recent time-series foundation models that have been pretrained on large-scale datasets from relevant domains, including weather, healthcare, energy, and environment. These include Chronos~\cite{ansari2024chronos}, Time-MoE~\cite{shi2024time}, TimesFM~\cite{das2024decoder}, Moirai~\cite{liu2024moirai}, Moment~\cite{goswami2024moment}, and Timer-XL~\cite{liu2024timer}, which offer strong generalization through large-scale pretraining. A comparison is given in Table~\ref{tab:tsfm_comparison}. To evaluate retrieval-augmented performance on diverse real-world domains, we integrate our retriever with three publicly available time-series foundation models: Time-MoE, Timer-XL, and Moment, which are selected based on the availability of stable, open-source implementations that support customization and downstream fine-tuning. We leave the adaptation of our retriever to additional proprietary or closed-source foundation models, as well as its integration into unified pretraining pipelines, for future work.

\xhdr{Comparison of \name with Time-series Foundation Models} It is important to note that our model is not itself a cross-domain foundation model, but rather a modular encoder-based retriever capable of enhancing such models. Architecturally, our model adopts an encoder-only design with flexible point- and patch-based tokenization, supports input sequences exceeding 2,048 tokens, and enables effective cross-channel interactions through channel-biased attention mechanisms.

\subsection{Experiment Configurations}\label{app:hyper}
All models are implemented in PyTorch and trained on NVIDIA A100 40GB GPUs. For most time series models, we adopt the implementation from TSLib \cite{TSLib}\footnote{\url{https://github.com/thuml/Time-Series-Library}}. The sequence length is fixed at 96 for both prediction horizons of 7 and 24. We use mean squared error (MSE) as the loss function for forecasting tasks, and accuracy for classification. Forecasting models are trained for 10 epochs, while classification models are trained for up to 150 epochs with early stopping. We follow the official code to implement other baselines~\cite{hu2025context}\footnote{\url{https://github.com/tokaka22/ICLR25-FSCA}}. All other hyperparameters follow the default settings in TSLib, except for those explicitly tuned to achieve the best performance, as reported in Tables~\ref{tab:baseline_hyper}. For our model, the initial learning rate is tuned from $\{10^{-4}, 10^{-3}\}$. The number of attention layers is tuned from $\{6,12\}$, and the hidden dimension is from $\{384, 768\}$ with the number of heads in $\{6,12\}$.

\begin{table}[htbp]\vspace{-0.5cm}
\centering
\caption{Best hyperparameters per model}\label{tab:baseline_hyper}
\resizebox{0.6\textwidth}{!}{
\begin{tabular}{lrrr}
\toprule
Model Name   & Learning Rate & Encoder Layers  & Hidden Dimension \\
\midrule
DLinear      & 0.0010& 2& 32\\
PatchTST     & 0.0050& 4& 64\\
TimeMixer    & 0.0100& 4& 64\\
TimesNet     & 0.0010& 4& 64\\
iTransformer & 0.0100& 4& 64\\
FSCA & 0.0001&4 & 256\\
\bottomrule
\end{tabular}}\vspace{-0.5cm}
\end{table}

\subsection{Embedding Visualization}\label{app:emb_vis}
\begin{figure}[htbp]
    \centering
    \begin{minipage}{0.5\linewidth}
        \centering
        \includegraphics[width=\linewidth]{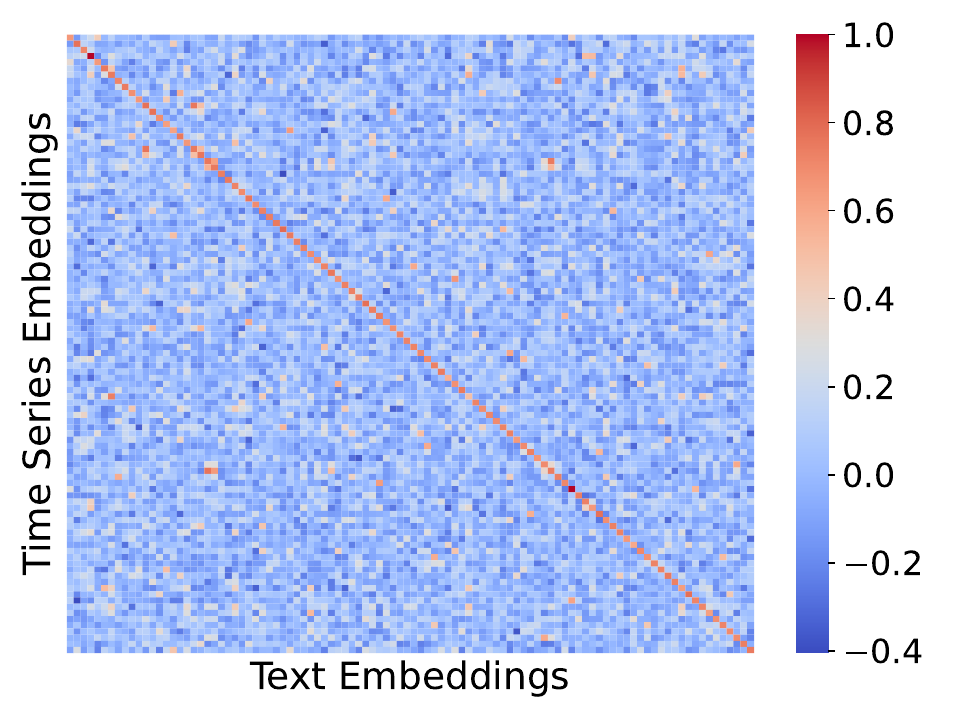}
        \caption{Cosine Similarity Matrix Between Text and Time Series Embeddings.}
        \label{fig:similarity}
    \end{minipage}\hfill
    \begin{minipage}{0.45\linewidth}
        \centering
        \includegraphics[width=\linewidth]{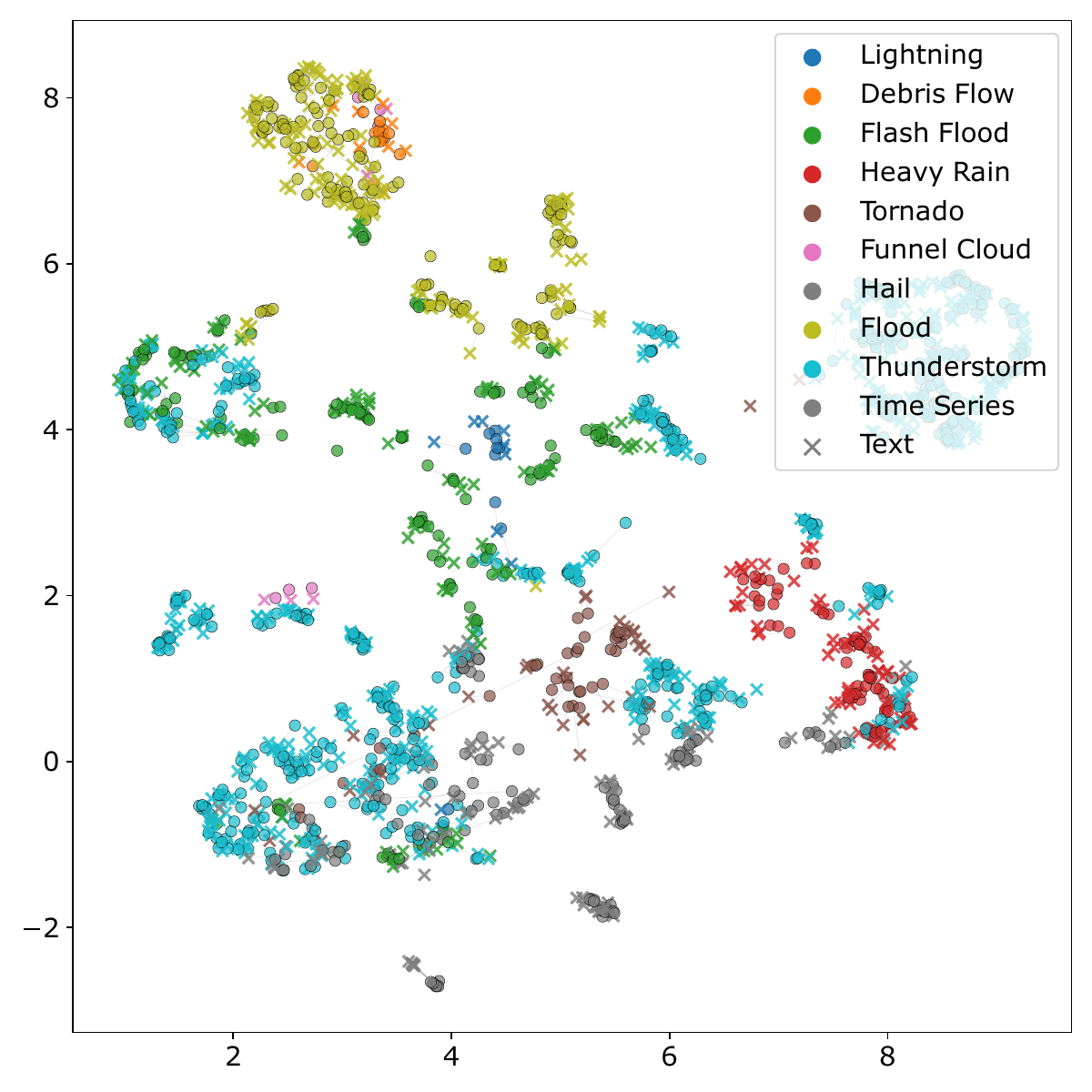}
        \caption{Umap Visualization of Aligned Text and Time Series Embeddings.}
        \label{fig:embedding_umap}
    \end{minipage}
\end{figure}

Figure~\ref{fig:similarity} presents the cosine similarity matrix between text and time series embeddings across the test set. The diagonal dominance indicates that \name successfully aligns each time series with its corresponding textual description, suggesting strong one-to-one semantic matching in the shared embedding space. Off-diagonal similarities remain low, demonstrating the model’s ability to distinguish unrelated instances. Figure~\ref{fig:embedding_umap} visualizes the joint embedding space using UMAP. Each color represents a distinct event category, where circles ($\circ$) denote time series instances and crosses ($\times$) denote their corresponding textual descriptions. A line connects each text–time series pair. We observe clear clustering by event type, with paired modalities positioned closely in the embedding space. Notably, for some events (\eg "Flood" and "Debris Flow"), clusters partially overlap, reflecting shared underlying dynamics. The tight alignment between paired points validates the effectiveness of our dual-level alignment strategy, and the modality-mixing within clusters suggests successful fusion of structured and unstructured signals.

\subsection{Classification Task}\label{app:classification_results}
\begin{wraptable}{r}{0.4\textwidth}
\centering\vspace{-0.5cm}
\caption{Weather Event Classification Accuracy and F1 Score (\%).}
\resizebox{\linewidth}{!}{
\begin{tabular}{llcc}
\toprule
\textbf{Model}&\textbf{Size} & \textbf{Accuracy}  & \textbf{F1} \\
\midrule
\multirow{2}{*}{Time-MoE} 
  & small  & 56.27 & 16.56 \\
  & large  & 59.09 & 19.74 \\
\midrule
\multirow{2}{*}{Moment} 
  & base   & 65.43 & 28.29 \\
  & large  & 64.94 & 26.35 \\
\midrule
\multirow{5}{*}{Chronos} 
  & tiny   & 74.79 & 40.21 \\
  & mini   & 73.89 & 37.98 \\
  & small  & 71.07 & 35.39 \\
  & base   & 71.42 & 36.40 \\
  & large  & 71.97 & 36.30 \\
\bottomrule
\end{tabular}}
\label{tab:full_classification_results}\vspace{-0.5cm}
\end{wraptable}
Table~\ref{tab:full_classification_results} reports the classification accuracy and F1 scores of different size variants of time-series foundation models on the weather event classification task. We observe that a larger model size does not necessarily lead to better performance. For example, Moment's base model achieves a higher F1 score than the large model despite a lower accuracy. In contrast, Chronos exhibits more stable performance across scales, with the tiny and mini variants achieving the best F1 scores, outperforming even the larger variants. These results suggest that, in domain-specific classification tasks with relatively limited supervision, scaling up foundation models may not always be beneficial, and smaller models can offer a better balance between accuracy and efficiency. 
\subsection{RAG Setting}\label{app:RAG}
In our retrieval-augmented generation (RAG) framework, given a query time series $\mathbf{X}_q$, we compute its \texttt{[CLS]} token embedding as $\mathbf{h}_q \in \mathbb{R}^d$ using the frozen encoder from \name. Based on cosine similarity, we retrieve the top-$R$ most relevant multimodal pairs ${(\mathbf{X}^i, \tau_{\text{cxt}}^i)}_{i=1}^R$ from the corpus, where $\mathbf{X}^i$ is a historical multivariate time series and $\tau_{\text{cxt}}^i$ is the associated sample-level context. Each retrieved pair is transformed into a soft prompt vector using a trainable linear projection layer. Specifically, the time series component is encoded to $\mathbf{h}_{\text{ts}}^{(i)} \in \mathbb{R}^d$, and the textual context $\tau_{\text{cxt}}^i$ is encoded to $\mathbf{z}{\text{cxt}}^{(i)} \in \mathbb{R}^d$ using a frozen SentenceTransformer, followed by a shared projection. For the \textit{TS+Text} setting, we concatenate each pair as $\mathbf{p}^{(i)} = [h_{\text{ts}}^{(i)} ; \mathbf{z}_{\text{cxt}}^{(i)}] \in \mathbb{R}^{2d}$, and stack all $R$ vectors to form the final prompt:
\begin{equation*}
\mathbf{P} = \texttt{Proj} \left( [\mathbf{p}^{(1)} ; \cdots ; \mathbf{p}^{(R)}] \right) \in \mathbb{R}^{d_f},
\end{equation*}
where \texttt{Proj} is a feedforward layer mapping from $\mathbb{R}^{2Rd} \rightarrow \mathbb{R}^{d_f}$, and $d_f$ is the hidden dimension of the downstream time series foundation model. For the \textit{TS-only} setting, we omit the text component and instead concatenate $[h_{\text{ts}}^{(1)}; \cdots; h_{\text{ts}}^{(R)}] \in \mathbb{R}^{Rd}$ and project into $\mathbb{R}^{d_f}$ accordingly.

This dense prompt $\mathbf{P}$ is prepended to the query sequence during inference. For decoder-only models (\eg Timer-XL, Time-MoE), $\mathbf{P}$ is appended to the autoregressive context at each decoding step. For encoder-only models (\eg Moment, \name), $\mathbf{P}$ is inserted as a prefix to the encoder input, \ie 
\begin{equation*}
    \hat{y}=\texttt{Head}([\mathbf{P}|\mathbf{H}_q]),
\end{equation*}
where $\mathbf{H}_q \in \mathbb{R}^{L \times d_f}$ is the encoded query and \texttt{Head} is a forecasting head trained from scratch. In all configurations, only \texttt{Proj} and \texttt{Head} are updated during training in RAG framework, while the backbone foundation model remains frozen.
\subsection{Standalone Time Series Encoder}\label{app:standalone_encoder}
To evaluate the classification capabilities of time series foundation models, we finetune a multi-layer perceptron (MLP) classifier on top of each model’s final output representation, as most existing time series foundation models do not support classification task by design, except Moment~\cite{goswami2024moment}, The MLP consists of four hidden layers with sizes $[256, 128, 64, 32]$, followed by a softmax output layer corresponding to 9 weather event categories. This architecture was selected based on empirical tuning for optimal performance on our classification task. We include all available variants from four foundation model families: Time-MoE, Timer-XL, Moment, and Chronos. All backbone parameters of the time series foundation models are fully activated and updated during training to ensure consistency and fair evaluation. Each model is finetuned for 100 epochs using the Adam optimizer. The training batch size is set to 256 for small and mid-sized variants, and reduced to 128 for larger models to accommodate memory constraints. 

For full-shot time series models, we train them from scratch using a unified training and evaluation protocol with time series foundation models. Results are shown in Table~\ref{tab:classification} and Table~\ref{tab:full_classification_results}.

\subsection{Empirical Case Study}\label{app:case_study}
\begin{figure}[htbp]
    \centering
    \includegraphics[width=0.95\linewidth]{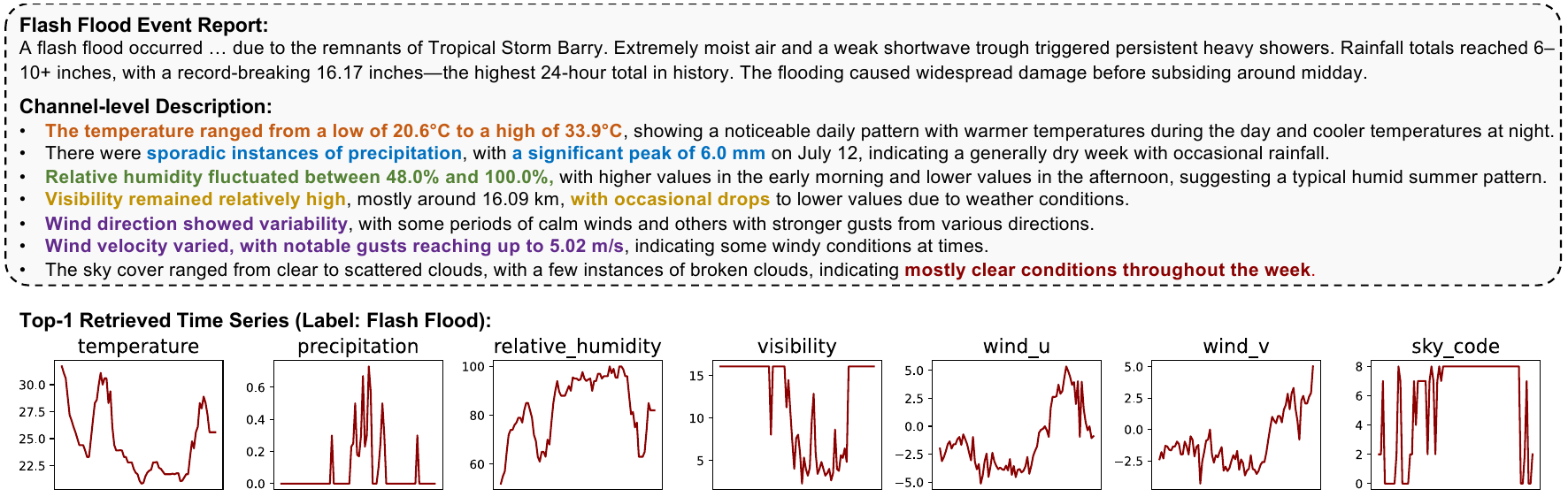}
    \caption{A case study of text-to-timeseries retrieval of flash flood-related time series. The key textual cues are highlighted in color for clarity.}
    \label{fig:text2ts_case_study}
\end{figure}

Figure~\ref{fig:text2ts_case_study} illustrates the capability to align detailed textual context with corresponding multivariate time series. The retrieval pool is constructed by excluding the query’s paired time series instance. This setup ensures that retrieved results are non-trivial and reflect the model's ability to identify semantically similar yet distinct examples. \name leverages both high-level and fine-grained semantic cues to retrieve the most relevant time series from the curated candidate pool. The top-1 retrieved sequence closely reflects key patterns in the query text, which can serve as a valuable reference for downstream forecasting, scenario simulation, or contextual explanation.

\subsection{Timeseries-to-Timeseries Retrieval}\label{app:ts2ts}

\begin{figure}[htbp]
    \centering
    \includegraphics[width=1\linewidth]{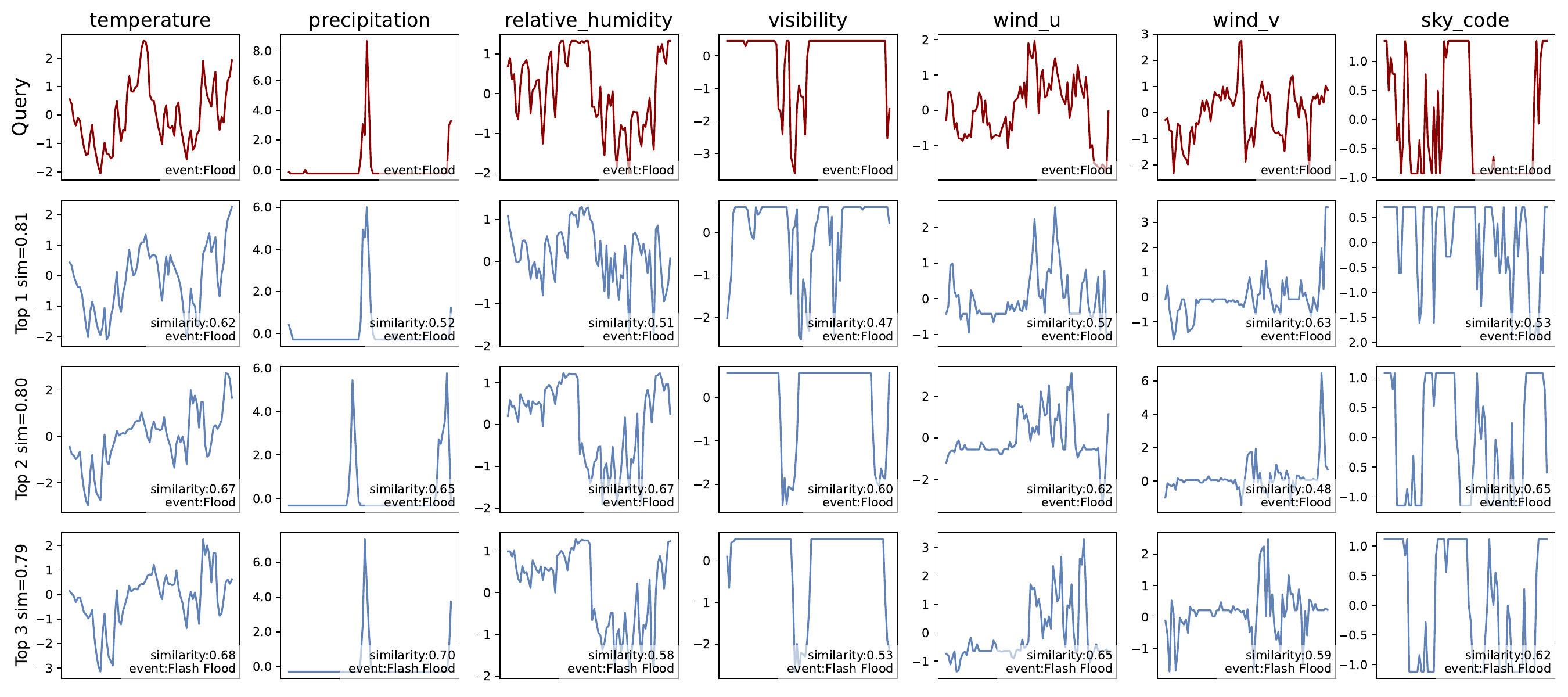}
    \caption{Visualization of Timeseries-to-Timeseries Retrieval by \name}
    \label{fig:ts2ts_retrieval_case}
\end{figure}
\xhdr{TS-to-TS Case Study} Figure~\ref{fig:ts2ts_retrieval_case} illustrates a case study of TS-to-TS retrieval using \name. Given a query time series (top row) labeled as \texttt{Flood}, the system retrieves the top-3 most similar samples from the corpus based on embedding similarity in the shared representation space. The similarity score for each retrieved sample is shown on the left, with per-channel similarity values annotated below each plot. We observe that all retrieved samples have high overall similarity scores (approximately 0.79–0.81), reflecting strong semantic alignment. The top-2 retrievals are also labeled as \texttt{Flood}, while the third belongs to a semantically related event, \texttt{Flash Flood}, suggesting that \name is capable of retrieving contextually relevant samples even across closely related labels. Notably, \name enables fine-grained channel-level similarity assessment by leveraging its Channel Identity Tokens (CIT), which allow independent embedding of channel-specific signals. 

However, we also find that high similarity in individual channels (\eg temperature or precipitation) does not always guarantee high overall semantic alignment. For instance, the first retrieval shows moderate similarity across channels but still achieves a high overall semantic score. This highlights the benefit of \texttt{TRACE}'s structured aggregation over all channels to capture global semantics and reveal the most semantically dominant channels that contribute most to the retrieval relevance. This capability enables \name to go beyond surface-level similarity, retrieving samples that share latent event signatures rather than merely matching patterns across all channels uniformly.
\subsection{Complexity and Efficiency}\label{app:efficiency}
\subsubsection{Computational Complexity}
We analyze the computational complexity of the main components in \name, including the encoder stage, the dual-level contrastive alignment, and the retrieval-augmented generation (RAG) setup.

\xhdr{1. Encoder Pre-training Complexity}
Let $X \in \mathbb{R}^{C \times T}$ be the input multivariate time series with $C$ channels and $T$ time steps. The sequence is tokenized into $\hat{T} = \lfloor T/P \rfloor$ patches per channel, each projected to a $d$-dimensional embedding. The total token length after flattening is $1+C(\hat{T}+1)$. This includes one global \texttt{[CLS]} token, one \texttt{[CIT]} token per channel, and $\hat{T}$ patch tokens per channel. The encoder is a $N$-layer Transformer with multi-head channel-biased attention. The complexity per attention layer is $\mathcal{O}(L^2d)=\mathcal{O}(C^2\hat{T}^2d)$. Note that channel-biased attention applies a sparse mask $M \in \{0,1\}^{L \times L}$ to restrict certain attention to within-channel interactions, which effectively reduces the constant factors in practice but not the asymptotic complexity.

\xhdr{2. Dual-level Contrastive Alignment}
Let $B$ be the batch size. For each time series, the alignment stage computes:
\begin{itemize}[leftmargin=5mm,itemsep=0.1pt,topsep=0.3pt]
    \item Sample-level similarity: $\mathcal{O}(B^2 d)$ for all $\mathbf{h}_{\texttt{[CLS]}}$–$\mathbf{z}_{\mathrm{cxt}}$ pairs.
    \item Channel-level similarity: For $C$ channels and $B$ instances, total cost is $\mathcal{O}(B^2 C^2 d)$ for $\mathbf{h}_c$–$\mathbf{z}_c$ pairs.
    \item Negative mining selects top-$R$ hardest negatives per instance and per channel, which costs $\mathcal{O}(B\log R+BC\log R)$, and is negligible compared to similarity computation.
\end{itemize}

\xhdr{3. Retrieval-Augmented Generation}
During inference, retrieval selects top-$R$ neighbors for a query based on cosine similarity:
\begin{itemize}[leftmargin=5mm,itemsep=0.1pt,topsep=0.3pt]
\item Retrieval cost: $\mathcal{O}(R d)$ using approximate methods (\eg FAISS) from a database.
\item Prompt generation: if soft prompt dimension is $d_f$, and each retrieved pair contributes $d$-dim vector, this yields a projection cost of $\mathcal{O}(Rdd_f)$.
\item The forecasting model remains frozen; only the soft prompt (a single vector of shape $[1, d_f]$) is appended, incurring no extra Transformer-layer cost.
\end{itemize}

\xhdr{Summary} Pre-training (Transformer encoder) yields $\mathcal{O}(L^2  d)$ per layer. Alignment yields $\mathcal{O}(B^2  d + B^2  C^2 d)$ and RAG inference yields $\mathcal{O}(R  d  d_f)$ for retrieval and projection.

\subsubsection{Empirical Runtime}
We report the model size and empirical runtime of \name and other baselines in Table~\ref{tab:runtime}, including FSCA~\cite{hu2025context}, which is the second-best train-from-scratch time series model, and time series foundation models with the availability of open-source implementations. \name activates only 0.12M parameters during finetuning with a lightweight linear head, which is nearly 200× fewer than FSCA and over 700× fewer than Time-MoE\textsubscript{small}. This lightweight design results in substantially faster training and inference speed. Compared to Moment, \name achieves faster training time with significantly fewer trainable parameters and better performance, which can be attributed to its multichannel modeling with channel-biased attention. While slightly slower than Timer-XL, which is a decoder-only model with causal attention, \name offers an acceptable overhead given its significantly stronger retrieval performance and the high quality of embeddings it produces for cross-modal and TS-to-TS retrieval.
It is worth noting that for Timer-XL and Time-MoE, despite their strong generalizability, parameter-efficient finetuning strategies are relatively underexplored, as all model parameters must be activated and updated during finetuning for reasonable performance in domain-specific tasks.
\begin{table}[htbp]
\centering\vspace{-0.5cm}
\caption{Comparisons of model efficiency. Activated Params indicates the number of parameters activated during finetuning for 7-step forecasting on the weather dataset. Training and inference time are seconds per epoch on the forecasting dataset. Device is a single A100 40GB GPU.}\label{tab:runtime}
\resizebox{0.8\linewidth}{!}{
\begin{tabular}{lccccc}
\toprule
& \textbf{Total Params} & \textbf{Activated Params} & \textbf{Training Time} & \textbf{Inference Time} \\ \midrule
\textbf{FSCA} &82.35M &22.68M &1249.701 &1.589  \\
\textbf{\name}&10.78M &0.12M &6.054&0.955\\
\midrule
\textbf{Moment}$_\text{base}$ &109.87M&0.24M&11.706&1.691\\
\textbf{Timer-XL}$_\text{base}$ &84.44M & 84.44M &3.392&0.685  \\
\textbf{Time-MoE}$_\text{small}$&113.49M&113.49M&106.308&15.545\\
\bottomrule
\end{tabular}}
\label{tab:efficiency}
\end{table}

\section{Discussion}\label{app:discussion}
\xhdr{Limitation}
While \name demonstrates strong performance in multimodal retrieval and retrieval-augmented forecasting, it currently assumes the availability of aligned time series–text pairs during training. In some domains, such alignment may be noisy or incomplete. Additionally, although channel-level alignment improves interpretability and fine-grained matching, it introduces a modest increase in computational overhead during training. We believe these trade-offs are justified by the performance gains but acknowledge that further optimization may enhance scalability.

\xhdr{Future Work}
In future work, we plan to extend \name to support weakly supervised and semi-supervised settings, where textual context is partially missing or noisy. Another promising direction is integrating domain adaptation techniques to improve generalization across unseen domains and sensor modalities (\eg image, video). Moreover, exploring autoregressive generation conditioned on retrieved time series–text pairs may further enhance understanding tasks in temporal modeling.

\xhdr{Broader Impact}
\name offers a general framework for cross-modal reasoning in time series applications, with potential benefits in domains such as healthcare monitoring, disaster forecasting, and industrial diagnostics. By improving retrieval and interpretation of structured temporal data, our approach may enhance decision support and model transparency. However, we encourage responsible deployment and emphasize the importance of auditing training data and retrieval outputs to avoid amplifying biases present in either modality.

\end{document}